\documentclass[11pt,twoside]{article} % For LaTeX2e

\usepackage[utf8]{inputenc} % allow utf-8 input
\usepackage[T1]{fontenc}    % use 8-bit T1 fonts
\usepackage{booktabs}       % professional-quality tables
\usepackage{amsfonts}       % blackboard math symbols
\usepackage{nicefrac}       % compact symbols for 1/2, etc.
\usepackage{microtype}      % microtypography

\usepackage{subfigure}
\usepackage{epsf}
\usepackage{epsfig}
\usepackage{fancyhdr}
\usepackage{graphics}
\usepackage{graphicx}
\usepackage{psfrag}
\usepackage{fullpage}
\usepackage{pdfpages}
\usepackage{tikz}
\usepackage{tikz-network}
\usepackage{bbm}
\usepackage{booktabs}
\usepackage{floatrow}
\usepackage{amsmath,amsfonts,bm}
\usepackage{url}% for url's in bib
\usepackage[colorlinks,linkcolor=magenta,citecolor=blue, pagebackref=true]{hyperref}
\renewcommand*{\backrefalt}[4]{%
    \ifcase #1 \footnotesize{(Not cited.)}%
    \or        \footnotesize{(Cited on page~#2.)}%
    \else      \footnotesize{(Cited on pages~#2.)}%
    \fi}
\usepackage{color}

\usepackage{amsthm}
\usepackage{amsmath}
\usepackage{amssymb,bbm}
\usepackage{caption}
\usepackage{algorithmic}
\usepackage{algorithm}
\usepackage{textcomp}
\usepackage{siunitx}
\usepackage{wrapfig}
\usepackage{algorithmic}
\usepackage{algorithm}
\usepackage{multirow}
\usepackage{multicol}% colors
\newtheorem{lemma}{Lemma}

\newtheorem{theorem}{Theorem}

\newtheorem{definition}{Definition}
\begin{document}

\begin{center}

{\bf{\LARGE{Improving Generative Flow Networks with Path Regularization}}}
  
\vspace*{.2in}
{\large{
\begin{tabular}{cccccc}
Anh Do$^{\ddagger, \star}$& Duy Dinh$^{\ddagger, \star}$& Tan Nguyen$^{\diamond, \star}$& Khuong Nguyen$^{\ddagger}$&Stanley Osher$^{\diamond, \star\star}$ &Nhat Ho$^{\dagger, \star\star}$ 
\end{tabular}
}}

\vspace*{.2in}

\begin{tabular}{c}
 University of Texas, Austin$^{\dagger}$; University of California, Los Angeles$^{\diamond}$; FPT Software AI Center$^{\ddagger}$
\end{tabular}

\today

\vspace*{.2in}

\begin{abstract}
Generative Flow Networks (GFlowNets) are recently proposed models for learning stochastic policies that generate compositional objects by sequences of actions with the probability proportional to a given reward function. The central problem of GFlowNets is to improve their exploration and generalization. In this work, we propose a novel path regularization method based on optimal transport theory that places prior constraints on the underlying structure of the GFlowNets. The prior is designed to help the GFlowNets better discover the latent structure of the target distribution or enhance its ability to explore the environment in the context of active learning. The path regularization controls the flow in GFlowNets to generate more diverse and novel candidates via maximizing the optimal transport distances between two forward policies or to improve the generalization via minimizing the optimal transport distances. In addition, we derive an efficient implementation of the regularization by finding its closed form solutions in specific cases and a meaningful upper bound that can be used as an approximation to minimize the regularization term. We empirically demonstrate the advantage of our path regularization on a wide range of tasks, including synthetic hypergrid environment modeling, discrete probabilistic modeling, and biological sequence design.
\end{abstract}

\end{center}
\let\thefootnote\relax\footnotetext{$^{\star}$ Anh Do and Duy Dinh contributed equally to this work. $^{\star\star}$ Nhat Ho and Stanley Osher are co-last authors of this work. Correspondence to: Nhat Ho (\href{mailto:minhnhat@utexas.edu}{minhnhat@utexas.edu}) and Tan Nguyen (\href{mailto:tanmnguyen89@ucla.edu}{tanmnguyen89@ucla.edu}).}

\section{Introduction}
\label{sec:introduction}
%Generative Flow Networks (GFlowNets)~\cite{DBLP:journals/corr/abs-2106-04399} is a new research direction in generative modeling, which allows neural networks to model distributions over data structures likes graphs, sets, sequences or other compositional objects constructed sequentially. A GFlowNet  is a trained stochastic policy or generative model for generating an compositional object from a sequence of actions (trajectory), such that the probability of generating an  object $\mathbf{x}$ is proportional to a given positive reward $R(\mathbf{x})$ for that object~\cite{DBLP:journals/corr/abs-2106-04399}. 
%By leveraging the generalization capability of Neural Networks to model the policy, GFlowNets can discover the latent structured of the data, then making them better to learn the distributions, extrapolate to new modes, and more robustness when dealing with out-of-distribution.
%Theoretically, ~\cite{https://doi.org/10.48550/arxiv.2111.09266} has shown that if the training objective such as flow-matching objective~\cite{DBLP:journals/corr/abs-2106-04399} or trajectory ballanced objective~\cite{Malkin2022TrajectoryBI}) is achieved on all states and possible trajectories respectively, then GFlowNets can be trained to completion, i.e., perfectly generating objects proportional to their rewards. %As a result, we could use any training policy, so long as it can sample all the possible trajectories with non-zero probability, i.e., GFlowNets training can be offline.

Recently proposed by Bengio et al.~\cite{DBLP:journals/corr/abs-2106-04399}, Generative Flow Networks (GFlowNets) are generative models for compositional objects, which learn a stochastic policy that sequentially modifies a temporarily constructed object through a sequence of actions to make the generating likelihood proportional to a given reward function. Specifically, GFlowNets aims to solve the problem of generating a diverse set of good candidates. In biological sequence design, diversity is a key consideration because of improving the chance of discovering candidates that can satisfy many evaluation criteria later in downstream phases. Especially, in the multi-round active learning setting, where the generator was iteratively improved by receiving feedback from an oracle on their proposed candidates, the effect of diverse generation becomes apparent because more diversity means more exploration and knowledge gained. Besides, the generalization ability of GFlowNets over structured data makes them a good framework for discrete probabilistic modeling.

The central problems of GFlowNets are improving exploration and generalization. In this work, we propose to \emph{train GFlowNets with an additional path regularization via optimal transport}~\cite{Villani2003TopicsIO}, which acts as a prior on the underlying structures of GFlowNets. The prior is designed to help the GFlowNets better discover the latent structure of the target distribution or enhance its ability to explore the environment in the context of active learning. Precisely, the path regularization via OT can help GFlowNets generate more diverse and novel candidates via maximizing OT distances between two forward policies or improving generalization via minimizing the OT distances.
%The main problem is that the numbers of possible trajectories are enormous so practically GFlowNets could not be trained to completion with limited visited trajectories.

\vspace{0.5 em}
\noindent
\textbf{For generalization:}
% To improve GFlownets' generalization, we propose the prior on the underlying structures of the GFlowNets by path regularization via Optimal Transport. Before explaining why Optimal Transport is the good choice for path regularization, we firstly mentioned about the our proposal prior as below
To improve GFlownets' generalization, we propose the following prior constraints: (i) Forward policies at nearby states along trajectories with high flow should be similar so they will prefer certain directions leading to high rewards. For example, in discrete image probabilistic modeling, the modes concentrate in a small region of space, so the policy must focus on some specific directions
; (ii) Trajectories related to positive objects (both have high rewards) must share their paths. As a result, the similarity of states along trajectories with high flow is higher than in other places. From a probabilistic perspective, we propose to measure the similarity of states $s$ and $s'$ in GFlowNets by the transition probability from $s$ to $s'$
; (iii) When a GFlowNet learns something, the flow of the GFlowNet must be sparse. Thus, our proposal priors encourage GFlowNets to refine their flow, i.e., enhance
flow on high flow trajectories and vice versa.
%pruning small flow, reduce redundancy and enhance flow on high flow trajectories.% Because GFlowNets can be train offline so that combining exploration training policy to discover new states with a sparsity prior to distillate knowledge and reduce redundancy helps GFlowNets improving their generalization.
%As we will see in section 3, the transition probability leading to a directed distances in GFlowNet, which not only the natural choice for distance in GFlowNet but also linking path regularization via OT to Entropy, Cross Entropy and an efficient implementation of the path regularization via optimal transport
    %\item một vài hướng nhất định , giống nhau, prefer action
    %\item When a GFlowNet learns something, the flow of the GFlowNet must be sparse, i.e., the entropies of either GFlowNet's policy or trajectory flow are small. 

\vspace{0.5 em}
\noindent
\textbf{For diversity and exploration: }
%In multi-round active learning setting, the GFlowNet generator can limitedly access to the true expensive oracle, so it is trained with a reward proxy, which is an approximation of the true oracle. Because the proxy itself is trained based on the feedback from the oracle about the GFlowNets' proposed samples, it is important to have diverse generated candidates to provide the proxy useful information. 
To encourage GFlowNets' policy to generate more diverse candidates, such as in the multi-round active learning setting, we propose to put a prior on two neighbor forward policies, such that the prior encourages the dissimilarity of the forward policies. For example, in biological sequence design, each action adds a fragment or an amino acid, and the dissimilarity prior makes the policies biased toward adding more distinct elements to a molecule, making the molecules more diverse.

%The central problem is that the numbers of possible trajectories are enormous so practically GFlowNets could not be trained to completion with limited visited trajectories. To tackles this problems, we propose to train GFlowNets with an additional path regularization via Optimal Transport (\citet{Villani2003TopicsIO}) as a prior on the underlying structures of the GFlowNets. 

%In this paper we present path regularization via Optimal Transport to improve GFlowNets’ generalization.
%In multi-round active learningsetting, the effect of diverse generation becomes apparent. The reason is that
%in this setting, the generator is a GFlowNet, which was trained with the reward function as a proxy learned based on the feedback of the oracle about the their proposed samples. The proxy could be seen as an approximation of the true oracle. The diverse generated candidate then provides more useful signals to train the proxy better.

\vspace{0.5 em}
\noindent
\textbf{Why optimal transport is a good solution?} First, for the similarity prior, we need a measure of “distance” between pairs of probability distributions. The Optimal Transport (OT) theory  (\cite{Villani2003TopicsIO}) studies how probabilistic mass can be optimally transported from the support of one
probabilistic distribution to others of another distribution given a cost function. The minimum transportation cost, called OT distance, can be used as a metric that quantifies the distance between two probability distributions. In the context of GFlowNets, we want the regularization effect on nearby states so we apply optimal transport distance for forward policies $P_F(\cdot|  s)$ and $P_F(\cdot|s')$ at two neighbor states $s$ and $s'$. In which, to compute the distance we solve an optimal transport problem from support points $u_i$ (children of $s$) to $v_j$ (children of $s'$) given a cost $c(u_i,v_j)$. The cost $c(u_i,v_j)$ is a distance between $u_i$ and $v_i$. Compared with KL divergence, the biggest problem of KL divergence is that it was infinite for a variety of distributions with unequal support. And in GFlowNet, the support of $P_F(\cdot|  s)$ and $P_F(\cdot|s')$ are usually not equal. As a result, minimizing the optimal transport distances encourages the similarity of the forward policies making GFlowNets prefer certain directions and improving the generalization, whereas maximizing the optimal transport distances between two forward policies makes GFlowNets generate more diverse and novel candidates. Moving to the second prior, we need a loss to measure the expectation of the transition probabilities from one state to another. In terms of probabilistic interpretation, we can rewrite the OT distance  as the minimum expectation of the transport cost
\begin{equation}
     d(\alpha, \beta) = \min_{\gamma \sim \Pi(\alpha,\beta)}{\mathbb{E}_{u,v\sim \gamma}{c(u,v)}}
\end{equation} 
where $\alpha: = P_F(\cdot|  s)$,  $\beta := P_F(\cdot|s')$ and $\Pi(\alpha,\beta)$ is the set of all joint distributions $\gamma(u, v)$ with marginals $\alpha(u), \beta(v)$. We propose to use $c(u,v) = \min_{\tau = u\to \dots \to v}{-\log(P(\tau|u))}$ which can be natural interpreted as the directed distance between two states in GFlowNet. Then minimizing optimal transport distance corresponds to maximizing the transition probability from children $u$ of $s$ to children $v$ of $s'$. Finally, the proposed cost $c(u,v)$ also links the path regularization to entropy and cross-entropy terms, which encourages sparsity of GFlowNets' flow when minimizing the optimal transport distance while encouraging exploration when maximizing the optimal transport distance.

    %the similarities of children $u$ of $s$ and children $v$ of $s'$ along the trajectories with high flow and the similarities of forward transition to the similar states.
    
    %To maximize transition probabilities from children of $s$ to children of $s′$. We propose to use the cost function as $c(u,v) = -\log(P(v|u))$
    
    %inverse transition probabilities from children of s to children of s′. Minimizing optimal transport distance is corresponding to maximize the similarities of children of s and children of s′ along the trajectories with high flow

    %Luckily, Optimal transport distance is the minimum expectation of the inverse transition probabilities from children of s to children of s.

    %and an efficient implementation of the path regularization via optimal transport
    
    %For the second prior, we propose to use $c(u,v) = \min_{\tau = u\to \dots \to v}{-\log(P(\tau|u))}$, the transition probabilities from a state to another, which can be natural interpreted as the directed distance between two states in GFlowNets. The proposed distance also links the path regularization via optimal transport to entropy, cross-entropy term. Which encourages sparsity of GFlowNets' flow when minimizing the optimal transport distance, while encourages exploration when maximizing the optimal transport distance.

\vspace{0.5 em}
\noindent
\textbf{Contributions.} In this work, we develop a novel path regularization to increase the mode diversity and generalization of the generative flow network. Our contributions
can be summarized as follows:
\begin{enumerate}
    \item We propose to train GFlowNets with an additional path regularization via optimal transport, which acts as a prior on the underlying structures of  GFlowNets to improve the generalization and exploration of GFlowNets.% We prove that optimal transport is the best choice for GFlowNets.
    \item We define a new directed distance between two states in GFlowNets. The distance not only is the natural choice for optimal transport distance in GFlowNet but also links the path regularization via optimal transport to entropy and cross-entropy.
    \item We derive an efficient implementation of the regularization by finding its closed-form solutions in specific cases and a meaningful upper bound that can be used as an approximation when we want to minimize the regularization term.
    %Which is used as transport cost to compute OT distance between forward policies at two neighbor state.
    %\item Moreover, choosing transport cost as ours propose directed distance between two state in GFlowNets leads to the upper bound for the path regularization via optimal transport. The upper bound then not only provides an efficient implementation
    %but also can explain the behaviors of the path regularization via entropy terms.
    %\item Under certain conditions, we derive a closed form solutions for  optimal transport distance. %with ours proposal transport cost. 
    % \item Beyond that, in general, we prove that we can sample a part of edges from trajectory $\tau$ to compute the path regularization
    
\end{enumerate}

\vspace{0.5 em}
\noindent
\textbf{Organization.} The paper is organized as follows. In Section \ref{sec:background}, we provide backgrounds on GFlowNets and optimal transport. In Section \ref{sec:path_reg}, we propose a new directed distance between two states in GFlowNets and then derive the formulation of path regularization via optimal transport. We will explain why this is the natural and optimal choice. Then we derive the upper bound and efficient implementation of the path regularization. We provide extensive experiment results of path regularization via Optimal Transport in Section \ref{sec:experiments} and conclude the paper with a few discussions in Section \ref{sec:conclusion}. Extra experimental and theoretical results and settings are deferred to the Appendices.
\section{Background}
\label{sec:background}
\subsection{GFlowNets} \label{subsec:gflownets}
 \textbf{Preliminaries} Given a \textit{compositional} space $\mathcal{X}$, where each object $x \in \mathcal{X}$ can be constructed by taking a sequence of discrete \textit{actions} from the action space $\mathcal{A}$.  Incrementally, the generation process modifies a temporarily constructed object, which is called a \textit{state} $s \in \mathcal{S}$. In addition, a special action indicates that the object is completely constructed and represents a \textit{terminal state}, such that $s=x \in \mathcal{X}$. These states and actions correspond to the vertices and edges of a directed acyclic graph $G=(\mathcal{S}, \mathcal{A})$. The construction of an object $x \in \mathcal{X}$ defines a \textit{complete trajectory}, which is a sequence of transitions $\tau=\left(s_{0} \rightarrow s_{1} \rightarrow \ldots \rightarrow s_{n} = x \rightarrow s_{f}\right)$ beginning from the \textit{initial state} $s_{0}$ and ending in the \textit{final state} $s_{f}$. Let $\mathcal{T}$ be the set of all complete trajectories. Without loss of generality, we may assume that each terminal state $s \in \mathcal{X}$ has only one outgoing edge, which is $s \rightarrow s_{f}$. This can be easily achieved by augmenting every terminal state $s$ with more outgoing edges by a new terminal state $s^{\text{T}}$ and a stop action $s \rightarrow s^{\text{T}}$.

%A \textit{complete trajectory} is a sequence of states $\tau = (s_0, s_1,...,s_n,s_f)$, where each transition $s_{t}\to s_{t+1}$ is an action in $\mathcal{A}$. For each transition $s\to s'\in \mathcal{A}$, we call $s$ is a parent of $s'$ and $s'$ is  a child of $s$. Let $\mathcal{T}$ be the set of complete trajectories.

\vspace{0.5 em}
\noindent
\textbf{Flows} A \textit{trajectory flow}~\cite{https://doi.org/10.48550/arxiv.2111.09266} is a nonnegative function $F: \mathcal{T} \mapsto \mathbb{R^+}$. The flow $F(\tau )$ can be interpreted as the probability mass associated with trajectory $\tau$. As an analogy with water flow, the flow $F(\tau )$ is the number of water particles sharing the same path $\tau$. 
As a results we can define the state flow via $F(s) = \sum_{s\in \tau}F(\tau)$, and the edge flow via $F(s\rightarrow s') = \sum_{ s \rightarrow s' \in \tau}F(\tau)$. We can associate a probability measure $P$ with the trajectory flow $F$. In which, there are two important conditional probabilities, the forward transition probabilities (forward policy) $P_{F}(s'|s):= F(s\to s')/F(s)$ is related to adding an element to build the objects, and the backward transition probabilities (backward policy) $P_{B}(s|s'):= F(s\to s')/F(s')$ is related to moving an element.

\vspace{0.5 em}
\noindent
\textbf{Learning GFlowNets} Theoretically, \cite{https://doi.org/10.48550/arxiv.2111.09266} has shown that if the training objective such as flow-matching objective~\cite{DBLP:journals/corr/abs-2106-04399} or trajectory balanced objective~\cite{Malkin2022TrajectoryBI}) is achieved on all states and possible trajectories respectively, then GFlowNets can be trained to completion, i.e., perfectly generating objects proportional to their rewards. In this paper, we use Trajectory Balance Objective because it brings more efficient credit assignment and faster convergence~\cite{Malkin2022TrajectoryBI}. We provide more background of GFLowNets in Appendix \ref{sec:appendix_background}. %As a result, we could use any training policy, so long as it can sample all the possible trajectories with non-zero probability, i.e., GFlowNets training can be offline.

\subsection{Optimal Transport Distance} \label{subsec:ot theory}
%\textbf{Transportation Plans and Joint Probabilities. } OT's modern formulation as a linear program is due to Kantorovich (\cite{Peyr2019ComputationalOT}): For two histograms $\textbf{a}$ and $\textbf{b}$ in the simplex $\Sigma_{n} \stackrel{\text { def }}{=}\left\{\mathbf{x} \in \mathbb{R}_{+}^{n}: \mathbf{x}^{\text{T}} \mathbbm{1}_{n}=1\right\}$ and  $\Sigma_{m} \stackrel{\text { def }}{=}\left\{\mathbf{y} \in \mathbb{R}_{+}^{m}: \mathbf{y}^{\text{T}} \mathbbm{1}_{m}=1\right\}$, we write $\mathbf{U}(\mathbf{a}, \mathbf{b})$ for the transportation polytope of $\mathbf{a}$ and $\mathbf{b}$:
%\begin{equation} \label{eq: admissible couplings}  
%    \begin{aligned}
       % \mathbf{U}(\mathbf{a}, \mathbf{b}) = \left \{ \mathbf{P} \in \mathbb{R}_{+}^{n\times m}: \mathbf{P}\mathbbm{1}_{m} = \mathbf{a}, \mathbf{P}^{\text{T}}\mathbbm{1}_{n} = \mathbf{b}\right \}
        % \end{aligned}
%\end{equation}
%The set of matrices $\mathbf{U}(\mathbf{a}, \mathbf{b})$ is bounded and identified by $n + m$ equality constraints, and therefore is a convex polytope (the convex hull of a finite set of matrices).

%Intro
%has established a natural and useful geometric tool for comparing measures supported on metric probability spaces. The theory 

\textbf{Transportation plans and joint probabilities}
For two discrete probability measures $\boldsymbol{\alpha}$ and $\boldsymbol{\beta}$ over some space $\mathcal{X}$, the admissible couplings set, which can be interpreted as the set of transportation plans or joint probability distributions, is defined as:
\begin{equation}
    \Pi\left(\boldsymbol{\alpha},\boldsymbol{\beta}\right) = \left \{ \pi \in \mathbb{R}_{+}^{k\times l}: \pi\mathbbm{1}_{l} =\boldsymbol{\alpha}, \pi^{\top} \mathbbm{1}_{k} = \boldsymbol{\beta}\right \}.
\end{equation}% where we can interpret the admissible couplings set as the set of transportation plans or joint probability distributions

%where $k = |\boldsymbol{\alpha}|$ and $l = |\boldsymbol{\beta}|$ denotes the number of supports of $\boldsymbol{\alpha}$ and  $\boldsymbol{\beta}$ respectively. . %For example, for two multinomial random variables $X$ and $Y$ taking values in $\{1, \cdots, k\}$ and $\{1, \cdots, l\}$, each with distribution $\boldsymbol{\alpha}$ and $\boldsymbol{\beta}$ respectively, each matrix $\pi \in \Pi\left(\boldsymbol{\alpha},\boldsymbol{\beta}\right)$ is a possible joint probabilities of $(X, Y)$, where $\pi_{ij} = P(X=i,Y=j)$.

%\textbf{Optimal transportation. } Given a $n \times m$ cost matrix $\mathbf{C}$, where $\mathbf{C}_{i j}$ describes the cost of transport mass from bin $i$ toward bin $j$, the cost of mapping $\textbf{a}$ to $\textbf{b}$ using a transportation plan (or joint probability) $\mathbf{P}$ can be defined as $\langle \mathbf{P}, \mathbf{C} \rangle$. The following problem:
%\begin{equation}
    %\begin{aligned}
%d_{\mathbf{C}}(\mathbf{a}, \mathbf{b}) \stackrel{\text { def }}{=} \min _{\mathbf{P} \in \mathbf{U}(\mathbf{a}, \mathbf{b})}\langle \mathbf{P}, \mathbf{C}\rangle
    %\end{aligned}
%\end{equation}
%\begin{equation}
    %\text{OT}_\mathbf{C}\left (\boldsymbol{\alpha},\boldsymbol{\beta} \right )
%\end{equation}
%is called an \textit{optimal transportation} problem between $\textbf{a}$ and $\textbf{b}$ given cost $\mathbf{C}$. 

\vspace{0.5 em}
\noindent
\textbf{Optimal transportation}
The Kantorovich optimal transport (\cite{Peyr2019ComputationalOT}) between $\boldsymbol{\alpha}$ and $\boldsymbol{\beta}$ is
 defined as follows:
 \begin{equation}
     \text{OT}_\mathbf{C}\left (\boldsymbol{\alpha},\boldsymbol{\beta} \right ) :=\min _{\pi \in \Pi\left(\boldsymbol{\alpha},\boldsymbol{\beta}\right)}\langle \mathbf{C},\pi\rangle
 \end{equation}
 where $\mathbf{C}$ is the cost matrix and $\mathbf{C}_{i j}$ describes the cost of transport mass from the support $i^{th}$ of $\boldsymbol{\alpha}$ toward the support $j^{th}$ of $\boldsymbol{\beta}$.  Whenever the matrix $\mathbf{C}$ is itself a metric matrix, the optimum of this problem, $ \text{OT}_\mathbf{C}\left (\boldsymbol{\alpha},\boldsymbol{\beta} \right )$, can be proved to be also a distance. Assuming that $k=l=d$, the worst-case complexity of computing that optimum with any of the algorithms known so far scales in $O\left(d^{3} \log d\right)$ and turns out to be super-cubic in practice (\cite{Pele2009FastAR}, $\S 2.1$).

%The cost usually is the Euclidean distance or other designed distances
%etween the supports two distribution.% Additionally, whenever the matrix $\mathbf{C}$ is itself a metric matrix, the optimum of this problem, $\text{OT}_\mathbf{C}\left (\boldsymbol{\alpha},\boldsymbol{\beta} \right )$, can be proved to be also a distance. %The network simplex method (\cite{Smith1994NetworkFT}, $\S9$) as well as other approaches (\cite{Orlin1988AFS}) can be used to get an optimal transport map $\pi^{\star}$ for this problem.
 
%Moreover, along this direction, there exists several works proposed efficient algorithms for solving the entropic OT (\cite{Altschuler2017NearlinearTA}, \cite{Lin2019OnEO}, \cite{Lin2019OnTA}) together with methods to stabilize these algorithms (\cite{Chizat2017ScalingAF}, \cite{Peyr2019ComputationalOT}, \cite{Schmitzer2019StabilizedSS}).

%The cost of computing optimal transport distance  scales at least in $\mathcal{O}(d^{3}log(d))$, where d is the number of support points (suppose $k=l=d$). By using the Sinkhorn algorithm  \cite{Cuturi2013SinkhornDL}, which solves optimal transport with entropic regularization, we can reduce computing costs to  $\mathcal{O}(d^{2})$. 

\section{Path Regularization via Optimal Transport}
\label{sec:path_reg}
%{\color{blue} Don't captitalize cross-entropy, entropy, optimal transport distance. Fix all of them. Reduce this Section to at most 4 pages.}
\subsection{Optimal Transport Formulation of the Path Regularization}
\par \textbf{Turn transition probability into directed distance}
We define a new directed distance between two states in GFlowNets, which is used as transport cost to compute OT distance.

\begin{definition}[ Directed Distance in GFlowNets] \label{Directed Distance}Let $\tau = s\to ... \to s'$ be  the transition from state $s$ to state $s'$ either via backward or forward policy.  The length of $\tau$ and the Directed Distance from $s$ to $s'$ are defined as follows: 
\begin{equation}
    \text{Len}(\tau) := -\log(P(\tau\mid s))
\end{equation}
\begin{equation}
     d(s,s') := \min_{\tau = (s\to ... \to s')}{-\log(P(\tau \mid s))}
\end{equation}
Where $d(s, s') = 0$ when $s \equiv s'$. Intuitively, $d(s,s')$ is the length of the shortage path from $s$ to $s'$.
\end{definition}
The proposed Directed Distance is the natural choice for distance in GFlowNets. Firstly, let consider two states $s$ and $s'$. They can be seen as the equivalent of one another, and have zero distance, if $P(s'|s)$, the probability of transitioning from state $s$ to state $s'$, is equal to $1$. In contrast, if the transition probability is equal to $0$, we cannot reach $s'$ from $s$ by following GFlowNets policy, then the distance must be infinite. Beside, it is obvious that $-\log(1)=0$ and $-\log(0) = \infty$. %Thus,  $\text{ len}(s\to s') := -\log(P(s'|s))$.
Another reason is that distances are additive, whereas probabilities are multiplicative. As a result, if we want the probability of a trajectory to be naturally related to its length, the transition probabilities between states should be changed to their distance by a "negative" logarithmic scale. 

\vspace{0.5 em}
\noindent
\textbf{Why do we use the proposed directed distance as transport cost to compute OT distance?} There are several reasons for using the directed distance as transport cost to compute OT distance. First, as discussed above, it is the natural choice to turn a transition probability into a distance. Second, with the transport cost as the directed distance, the OT distance can be decomposed into cross-entropy, entropy, and other negative terms. This gives us an upper bound for optimal transport distance and imposes sparsity prior for GFlowNets' structures when minimizing the OT distance or exploration when maximizing the OT distance. Besides, the choosing of cost function makes minimizing optimal transport distance correspond to maximizing the transition probability between children of neighbor states. Using the directed distance as transport cost allows us to derive closed-form solutions for optimal transport distance under some conditions. In contrast, if we use optimal transport distance with another transport cost, such as the shortage path in the graph, we only archive the first prior.

\vspace{0.5 em}
\noindent
\textbf{Optimal transport formulation of the path regularization:}
Once we have defined the directed distance between two states in GFlowNets, we can define the optimal transport distance between two forward policies of neighbor states. Consider two neighbor states $s < s'$ in trajectory $\tau$. The forward policy $P_F(\cdot| s)$ is a discrete probability measure supported by $\text{Child}(s) =\left\{ u_{1},...,u_{m} \right\}$  and $P_F(\cdot|  s')$ is a discrete probability measure supported by $\text{Child}(s') =\left\{ v_{1},...,v_{m} \right\}$.

%$$\mathbf{p_s} =\left\{ p_s^i =P_F(u_{i}\mid s) \right \} \qquad \mathbf{p_{s'}} =\left\{ p_{s'}^j =P_F(v_{j}\mid s') \right \}$$
The optimal transport (OT) distance between $P_F(\cdot|  s)$ and $P_F(\cdot|s')$ can be define as:
\begin{equation} \label{eq: OT distance}
    \text{OT}_\mathbf{C}\left ( P_F(\cdot| s),P_F(\cdot| s') \right ) :=\min _{\pi \in \prod\left(P_F(\cdot| s),P_F(\cdot| s')\right)}\langle \mathbf{C},\pi\rangle
\end{equation}
where the admissible couplings set is defined as:
\begin{equation} \label{eq: constraint coupling}
    \begin{aligned}
        \Pi(P_{F}(\cdot|s), P_{F}(\cdot|s')) : = \left \{ \pi \in \mathbb{R}_{+}^{k\times l}: \pi\mathbbm{1}_{l} = P_{F}(\cdot|s), \pi^{\top}\mathbbm{1}_{k} = P_{F}(\cdot|s')\right \}
    \end{aligned}
\end{equation}
and $\mathbf{C}$ is a cost matrix whose each entry is the length of a shortage path from $u_{i}$ to $v_{j}$ 
\begin{equation}
    \mathbf{C}_{ij} = c(u_{i}, v_{j}) := d(u_i,v_j) = \min_{\tau = (u_i\to ... \to v_j)}{-\log(P(\tau\mid u_i))}.
\end{equation}
Because we cannot access the full graph during training progress so the shortage path from $u_i$ to $v_i$ can only be constructed from the sub-graph containing $s_{t}$, $s_{t+1}$, their children states, and the edges that connecting them. Rather than going directly from $u_{i}$ to $v_{j}$ if this edge exists, we can always move from $u_i$ to $v_j$ along a back-and-forth trajectory, i.e., $u_{i} \rightarrow s\rightarrow s' \rightarrow v_{j}$, with the probability $P_{B}(s | u_{i})P_{F}(s' | s)P_{F}(v_{j} | s')$. Because transport cost from $u_i$ to $v_i$ is the length of the shortage path from $u_i$ to $v_i$, it is reasonable to have a minimum operator in the following formulation:
\begin{equation} \label{eq: transport cost}
    \begin{aligned}
        \mathbf{C}_{ij} = \left\{\begin{matrix}
0, & \text{if} \quad u_{i}\equiv v_{j} \\ 
\min\left(-\log(P_{B}(s \mid u_{i})P_{F}(s' \mid s)P_{F}(v_{j} \mid s')),-\log(P(v_{j}\mid u_{i}))\right), & \text{else if} \quad  u_{i} \rightarrow v_{j} \in \mathcal{A} \\ 
-\log(P_{B}(s \mid u_{i})P_{F}(s' \mid s)P_{F}(v_{j} \mid s')), & \text{otherwise.} 
\end{matrix}\right.
    \end{aligned}
\end{equation}
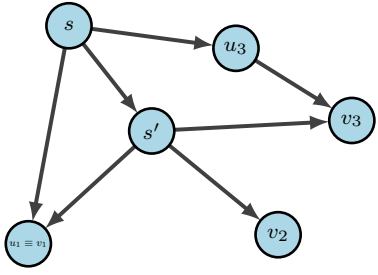
\begin{figure}[h] 
\floatbox[{\capbeside\thisfloatsetup{capbesideposition={right,top}, capbesidewidth=8.5cm}}]{figure}[\FBwidth]{
	\centering
	\begin{tikzpicture}
	\Vertex[x=0,y=0, label=$s$]{S}
	\Vertex[x=-0.54,y=-2.9,label=$u_1\equiv v_1$, fontscale = 0.5]{U1}
	\Vertex[x=2.22,y=-0.3, label=$u_3$]{U3}
	\Vertex[x=1.1,y=-1.4, label=$s'$]{Sp}
	\Vertex[x=2.78,y=-2.78,label=$v_2$]{V2}
	\Vertex[x=3.76, y=-1.26, label=$v_3$]{V3}

	\Edge[Direct](S)(U1)
	\Edge[Direct](S)(Sp)
	\Edge[Direct](S)(U3)
	\Edge[Direct](Sp)(U1)
	\Edge[Direct](Sp)(V2)
	\Edge[Direct](Sp)(V3)
	\Edge[Direct](U3)(V3)
	\end{tikzpicture}
	}
% 	\caption{\tiny{Optimal transport distance between forward policies $P_F(\cdot|s)$ and $P_F(\cdot|s')$. The forward policy $P_F(\cdot| s)$ is a discrete probability measure supported by $\text{Child}(s) =\left\{ u_{1},u_2 := s',u_{3} \right\}$. Similarly, the forward policy $P_F(\cdot| s')$ is a discrete probability measure supported by $\text{Child}(s') =\left\{ v_{1} \equiv u_1 ,v_2,v_{3} \right\}$. The cost matrix is a $3\times 3$ matrix. For example, $c_{11} = d(u_1,v_1) = 0$ (because $u_1\equiv v_1$). There are many possible paths to move from $u_3$ to $v_3$. First, going directly from $u_{3}$ to $v_{3}$ with a distance is equal to $\text{len}(u_3\rightarrow v_3) = -\log(P_F(v_3|u_3))$. Second, we can move from $u_3$ to $v_3$ along a back-and-forth trajectory, i.e., $u_{3} \rightarrow s\rightarrow s' \rightarrow v_{3}$, with a distance $-\log(P_{B}(s | u_{3}))-\log(P_{F}(s' | s)P_{F})-\log(P(v_{3} | s')))$. Because transport cost from $u_3$ to $v_3$ is the length of the shortage path from $u_3$ to $v_3$, $c_{33}=d(u_3,v_3) = \min(-\log(P_{B}(s | u_{3})P_{F}(s' | s)P_{F}(v_{3} | s')),-\log(P_F(v_3|u_3)) )$.}} 
{\caption{\small{Optimal transport distance between $P_F(\cdot|s)$ and $P_F(\cdot|s')$. The forward policy $P_F(\cdot| s)$ is a discrete probability measure supported by $\text{Child}(s) =\left\{ u_{1},u_2 := s',u_{3} \right\}$. Similarly, $P_F(\cdot| s')$ is supported by $\text{Child}(s') =\left\{ v_{1} \equiv u_1 ,v_2,v_{3} \right\}$. The cost matrix is a $3\times 3$ matrix. For example, $c_{11} = d(u_1,v_1) = 0$ (because $u_1\equiv v_1$). There are many possible paths to move from $u_3$ to $v_3$. First, going directly from $u_{3}$ to $v_{3}$ with a distance is equal to $\text{len}(u_3\rightarrow v_3) = -\log(P_F(v_3|u_3))$. Second, we can move from $u_3$ to $v_3$ along a back-and-forth trajectory, i.e., $u_{3} \rightarrow s\rightarrow s' \rightarrow v_{3}$, with a distance $-\log(P_{B}(s | u_{3}))-\log(P_{F}(s' | s)P_{F})-\log(P(v_{3} | s')))$. Because transport cost from $u_3$ to $v_3$ is the length of the shortage path from $u_3$ to $v_3$, $c_{33}=d(u_3,v_3) = \min(-\log(P_{B}(s | u_{3})P_{F}(s' | s)P_{F}(v_{3} | s')),-\log(P_F(v_3|u_3)) )$.}}}
	\label{Illustration OT}
\end{figure}
\begin{definition}[Optimal Transport Formulation of the Path Regularization] \label{def: Path Regularization}
For any complete trajectory $\tau = (s_{0}\rightarrow s_{1} \rightarrow ... \rightarrow s_{n})$, we define the path regularization via optimal transport as follows:
\begin{equation}\label{def: eq path regularization}
    \begin{aligned}
        \mathcal{L}_{\text{OT}}(\tau ) := \sum_{t=0}^{n-1}{ \text{OT}_{\mathbf{C}_{t;\theta}}\left ( P_F(\cdot| s_t;\theta),P_F(\cdot| s_{t+1};\theta) \right ) }
    \end{aligned}
\end{equation}
where $\mathbf{C}_{t;\theta}$ is the directed distance in GFlowNets (see Definition~\ref{Directed Distance}), with the formulation in Eqn.~\eqref{eq: transport cost}.
\end{definition}

If $\overline{\pi}_\theta$ is a training policy – usually that given by $P_F(\cdot| \cdot;\theta)$ or a tempered version of it – then the trajectory loss is updated along trajectories sampled from $\overline{\pi}_\theta$, i.e., with stochastic gradient:
\begin{align}
        \mathbb{E}_{\tau \sim \overline{\pi}_\theta}\nabla_{\theta}(\mathcal{L}_{\text{TB}}(\tau)+ \lambda \mathcal{L}_{\text{OT}}(\tau)).
\end{align}
%In definition \ref{def: Path Regularization},the subscript $\theta$ emphasize that the cost matrix $\mathbf{C}_{t;\theta}$ depend on $\theta$ because it is computed based on the Backward and Forward policies at $s_t$ and $s_{t+1}$. Therefore, the gradient of $\mathcal{L}_{\text{OT}}(\tau )$ not only affect the Forward policies at $s_t$ and $s_{t+1}$ but also the cost matrix, i.e, the transition probabilities between children of $s_t$ and $s_{t+1}$. For simplicity, We may omit the cost matrix and the parameters $\theta$, as in $\text{OT}\left ( P_F(\cdot| s_t),P_F(\cdot| s_{t+1}) \right )$.

\vspace{0.5 em}
\noindent
\textbf{The effects of minimizing the path regularization:} Remind that the optimal transport (OT) distance between $P_F(\cdot|  s)$ and $P_F(\cdot|s')$ is defined as follows:
\begin{equation}
    \text{OT}_\mathbf{C}\left ( P_F(\cdot| s),P_F(\cdot| s') \right ) :=\min _{\pi \in \prod\left(P_F(\cdot| s),P_F(\cdot| s')\right)}\langle \mathbf{C},\pi\rangle.
\end{equation}
Minimizing $\text{OT}_\mathbf{C}\left ( P_F(\cdot| s),P_F(\cdot| s') \right )$ makes $P_F(v|s')$ closed to $P_F(u|s)$, where $c(u,v)$ small. This is because all probability mass form $u$ can be transferred to $v$ with a smaller cost than to other places, so the total transport cost reduces. Also, $P_F(v|s')$ and $P_F(u|s)$ will be increased, where $c(u,v)$ small. This induces the similarity of the forward policies and makes the flow focus on specific directions. It should be noticed that the matrix cost $\mathbf{C}$ is not a constant but depends on the forward policies. Therefore, minimizing the optimal transport distance affects not only the forward policies but also the cost. However, the effect on the forward policies is also consistent with the effects on the cost. As in Eqn.~\eqref{theorem: eq closeform} of Theorem~\ref{theorem: close form} increasing $P_F(u|s)$ leads to increase $P^{*}_{B}(u| s)$. Besides, $-\log(P_{B}(s |u)P_{F}(s' | s)P_{F}(v | s'))$ is an upper bound of $c(u,v)$, so increasing $P_F(u| s)$ and $P_F(v|s')$ also make $c(u,v)$ smaller. We have $\mathbf{C}\geq 0$ so $\text{OT}_\mathbf{C}\left ( P_F(\cdot| s),P_F(\cdot| s') \right ) \geq 0$. Also, the optimal transport distance is equal to zero when all probability mass $P_F(v|s') = P_F(u|s)$ concentrate on $u,v$ where $c(u,v)=0$. For examples,  $P_F(s_t|s_{t-1}) = 1$ and $P_F(s_{t+1}|s_{t})=1$ is a special case. When GFlowNets visit a high reward object, the prior by designed helps GFlowNets quickly adapt their flow to this high reward terminal state. More discussion about entropy and cross-entropy is in Theorem \ref{thoerem: Upperbound}.

\vspace{0.5 em}
\noindent
\textbf{The effects of maximizing the path regularization:}
In contrast to minimizing the optimal transport distance, maximizing makes the forward policies different, so more diverse actions are chosen, leading to more diverse and novel candidates. As in the Theorem~\ref{thoerem: Upperbound}, the upper-bound of the optimal transport distance is the entropy of forward policy $\mathbf{H}(P_{F}(\cdot| s))$ and the entropy of the path $\mathbf{H}(P(\tau))$. Thus, maximizing the OT distance means maximizing the upper bound, which increases the entropy. This means more diversity and exploration. Remind that
in terms of probabilistic interpretation, we can rewrite the OT distance  as the minimum expectation of the transport cost
\begin{equation}
    \min_{\gamma \sim \Pi(\alpha,\beta)}{\mathbb{E}_{u,v\sim \gamma}{c(u,v)}}.
\end{equation}
Thus, maximizing the OT distance means maximizing the cost $c(u,v)$. Because $c(u,v)$ is an inverse function of transition probability from $u$ to $v$, maximizing the OT distance means minimizing $P(u\to v)$. As a result, the flow was distributed to more states, so more diverse action was chosen. %We have proved empirically that  path regularization via optimal transport is better than normal entropy regularization. One of the reasons is that our path regularization using the information from the structure of the DAG graph of GFlowNets and affect both backward and forward policies. While entropy regularization only focus on the forward policy at each state separately, our path regularization compares two neighbor states.

\subsection{Upper Bound and Efficient Implementation of the Path Regularization}

The cost of computing optimal transport distance $\text{OT}\left ( P_F(\cdot| s),P_F(\cdot| s') \right ) $ scales at least in $\mathcal{O}(d^{3}log(d))$, where d is the number of support points. By using the Sinkhorn algorithm~\cite{Cuturi2013SinkhornDL}, which solves optimal transport with entropic regularization, we can reduce computing costs to  $\mathcal{O}(d^{2})$~\cite{altschuler2017near, lin2019efficient, Lin-2022-Efficiency}. However, the implementation of the path regularization requires the computation of the optimal transport distances for all edges in the trajectory $\tau$. This is a significant burden on computing resources and capacity. To overcome this problem, in Theorem \ref{thoerem: Upperbound} we propose an upper bound of optimal transport distance $\text{OT}\left ( P_F(\cdot| s),P_F(\cdot| s') \right )$. The upper bound not only provides an efficient implementation but also can explain the behaviors of the path regularization. This is because the optimal transport distance can be decomposed into cross-entropy, entropy, and other negative terms. These negative terms are the total cost of an optimal transport problem with a negative sparse cost matrix, which can be solved completely in the hypergrid environment and EB-GFN experiments or other GflowNet settings under certain conditions (Section \ref{subsec: close form}). %Beyond that, in general, we prove that we can sample a part of edges from $\tau$ to compute the path regularization.

\begin{definition}[Pseudo backward policy]
Let $s$ be a state on trajectory $\tau$, which has its children set: $\text{Child}(s)=\{u_{1},...,u_{k}\}$, then the "pseudo backward policy" at state $s$ is defined as:
\begin{equation}
P^{*}_{B}(\cdot| s)) = (P^{*}_{B}(u_{1}|s), \dots, P^{*}_{B}(u_{k}| s))
\end{equation}
\textit{where we have}
\begin{equation} \label{eq: pseudo backward}
    P^{*}_{B}(u_{i}| s) = P_{B}(s| u_{i})  \quad \forall 1 \leq i \leq k.
\end{equation}

\end{definition}

\begin{theorem}[Upper bound of optimal transport distance] For any trajectory $\tau = (s_{0}\rightarrow s_{1} \rightarrow ... \rightarrow s_{n})$. The path regularization via optimal transport \label{thoerem: Upperbound} can be upper-bound by:
\begin{equation}
    \mathcal{L}_{\text{OT}}(\tau ) \leq \mathcal{L}_{\text{UB}}(\tau),
\end{equation}
\textit{where:}
\begin{equation}
        \mathcal{L}_{\text{UB}}(\tau ) := \sum _{t=0}^{n-1}\left [  \mathbf{H}(P_{F}(\cdot| s_{t}),P^{*}_{B}(\cdot| s_{t})) - \log(P_{F}(s_{t+1}|s_{t})) + \mathbf{H}(P_{F}(\cdot| s_{t+1})) \right].
\end{equation}
\end{theorem}
The proof of Theorem~\ref{thoerem: Upperbound} is in Appendix \ref{proof: upperbound}. The cross-entropy and entropy terms in the upper-bound $\mathcal{L}_{\text{UB}}(\tau )$ encourage the sparsity of GFlowNets’ flow. When
we minimizes the upper-bound $\mathcal{L}_{\text{UB}}(\tau )$, the cross-entropy terms $\mathbf{H}(P_{F}(.| s_{t}),P^{*}_{B}(.| s_{t}))$ try to match the Forward policy and the Pseudo backward policy at $s_t$. For example, consider a children state $u$ of $s_t$, if $P_{F}(u| s_{t})$ is high, then $P_{B}(s_t|u)$ is also high. As a result, increasing $P_{B}(s_t|u)$ makes other flows leading to $u$ pruned. While the meanings of cross-entropy and entropy terms are clear, we now would like to explain the meaning of regularization terms $- \log(P_F(s_{t+1}|s_t))$. Direct calculations indicate that:
\begin{align}
    \sum_{t=0}^{n-1}{- \log(P_F(s_{t+1}|s_{t}))} & =-\log \left(\prod_{t=0}^{n-1}{P_F(s_{t+1}|s_{t})}\right) = -\log(P(\tau)), \\
    \mathbb{E}_{\tau \sim \pi _{\theta}}(-\log(P(\tau))) & = \mathbf{H}(P(\tau)).
\end{align}
By taking this approach, the upper-bound not only regularizes the policies via $\mathbf{H}(P_{F}(\cdot| s_{t+1}))$ but also more importantly the path via $\mathbf{H}(P(\tau))$. Besides, we can think $-\log(P(\tau)) = -\log(P(\tau|s_0).P(s_0))=-\log(P(\tau|s_0))$ as the length of $\tau$. Thus the effect of minimizing path regularization also is biasing to shorter paths in terms of the direct distance and equivalently to the small numbers of paths with high flows. %Also, the effect of minimizing path generalization is diversity and exploration.

\subsection{Closed form formulation for the Path Regularization}\label{subsec: close form}
In the case of the hypergrid environment~\cite{DBLP:journals/corr/abs-2106-04399} and EB-GFN experiments~\cite{pmlr-v162-zhang22v}, Protein and DNA design or other GflowNet settings with certain
conditions, Theorem~\ref{theorem: close form} allows us to compute optimal transport loss from a closed form formulation rather than using the Sinkhorn Algorithm. The closed form formulation was derived by taking advantage of the following observations. For the first observation, any two neighbor states $s < s'$ do not have the same child state. The reason is that there is not exist an action $a_k$ such that $a_k$ can decompose into others actions, i.e, $a_k = a_i+a_j$. For the second observation, there is an action $a^*$ moving from a child $u\neq s'$ of $s$ to a child $v$ of $s'$, if and only if these children are achieved by taking the same action both at $s$ and $s'$. Moreover, $a^*$ is the action moving from $s$ to $s'$. Without loss of generality, we can denote $a_i$ as an action so that the state-action pair $(s,a_i)$ leads to $u_i$ and $(s',a_i)$ leads to $v_i$. Let denote $A^*_s$ as the set of valid non-terminal actions at state $s$. If the state-action pair (s, a) leads to s', and the state-action pair (s', a') leads to s", Let denote $s' = s+a$ and $s" = s+a+a'$. Moreover, if exist an action $a^*$ so that $s"=s+a^*$ then we denote $a^* =  a+a'$.

%Without loss of generality, we can denote $a_i$ as an action so that the state-action pair $(s,a_i)$ leads to $u_i$ and $(s',a_i)$ leads to $v_i$. For example,  in Hyper-grid environment (\cite{DBLP:journals/corr/abs-2106-04399}), $a_i$ is the action increasing coordinate $i$ by 1. Besides, for every nonterminal state $s$, there is also a termination action $a^\top$ that transitions to acorresponding terminal state $s^\top$. In EB-GFN experiments (\cite{DBLP:journals/corr/abs-2106-04399}), $a_{2i}$ is the action assigning pixel $i$ to 0 and $a_{2i+1}$ is the action assigning pixel $i$ to 1. In the environments of the above experiments, the size of the action space for a state $s$ is not a constant. For example, the actions exit the grid and the actions assign $0$ or $1$ to painted pixels are invalid. For convenience, if action $a_i$ is invalid at state $s$, we assign $P_F(u_{i}\mid s) := 0$, so the cost matrix of the optimal transport distance still is a square matrix with the zero cost according to invalid actions. Let us denote $A^*_s$ as the set of nonterminal valid actions at state $s$. If the state-action pair (s, a) leads to s', and the state-action pair (s', a') leads to s", Let denote $s' = s+a$ and $s" = s+a+a'$. Moreover, if exist an action $a^*$ so that $s"=s+a^*$ then we denote $a^* =  a+a'$.

%In the case of Grid and EB-GFN experiments, we can compute optimal transport distance for non-terminal states $s,s'$ from the following formulation rather than using the Sinkhorn Algorithm.

\begin{theorem}[Closed form solution for optimal transport distance] \label{theorem: close form} For any non-terminal neighbor states $s < s'$, let denote $a_i$ as an action so that the state-action pair $(s,a_i)$ leads to $u_i$ and $(s',a_i)$ leads to $v_i$, where $u_i \in \text{Child}(s)$ and $v_i \in \text{Child}(s')$. Let $A^*_s$ as the set of non-terminal valid actions at state $s$. Let  $a^*_s$ is the action moving from $s$ to $s'$. If $a_i \neq a_k+a_h \quad \forall a_i,a_k,a_h \in \mathcal{A}
$ and $a_i+a_h = a_m+a_n, a_i \neq a_m \Longleftrightarrow a_i = a_n , a_h = a_m, a_i \neq a_m$ then:
\begin{equation}\label{theorem: eq closeform}
\begin{aligned}
    \text{OT}\left ( P_F(\cdot| s),P_F(\cdot| s') \right ) &= \mathbf{H}(P_{F}(\cdot| s),P^{*}_{B}(\cdot| s)) - \log(P_{F}(s'|s)) + \mathbf{H}(P_{F}(\cdot| s'))\\&+P_F(s'|s).(\log( P_B(s'|s))+\log(P_F(s'|s)))
    \\&+\sum_{i\in A^{*}_s\bigcap A^{*}_{s'} }{\min(P_F(u_i|s),P_F(v_i|s'))c'_{i}},
\end{aligned}
\end{equation}
\begin{equation}
    u_i \neq s' \Longrightarrow u_i + a^*_s = v_i.
\end{equation}
where we define
\begin{equation}\label{eq: def cprime}
c'_{i}= 
\begin{cases}
  \min(0,\log( P_B(s|u_i))+\log(P_F(s'|s))+\log(P_F(v_i|s')) - \log(P_F(v_i|u_i))), & \text{if $u_i\neq s'$} \\
  0, & \text{if $u_i = s'$}
\end{cases}
\end{equation}
\end{theorem}

The proof of Theorem \ref{theorem: close form} and the closed form solution for optimal transport distance at terminal state are in Appendix \ref{proof: close form}. These close forms of the optimal transport distance will be used in our experiments in Section~\ref{sec:experiments}
(see Appendix \ref{proof: close form} for the reasons why we can apply Theorem~\ref{theorem: close form}).

%i.e,
%\begin{equation}
%     a_i \neq a_k+a_h \quad \forall a_i,a_k,a_h \in \mathcal{A} \Longleftrightarrow Child(s)\cap Child(s') = \O 
%\end{equation}
%and unique factorization, i.e,
%\begin{equation}
%    a_i+a_h = a_m+a_n \Leftrightarrow (a_i,a_h) = (a_m,a_n) \text{ or } (a_i,a_h) = (a_n,a_m) \Longleftrightarrow 
%    v_j \in Child(u_i)\Leftrightarrow a_i=a_j\neq a^\top  
%\end{equation}

\section{Experimental Results}
\label{sec:experiments}
%{\color{blue} Reduce this Section to at most 2 pages. Only need to keep one experiment in the main text while moving the rest to Appendix.}
In this section, we numerically justify the advantage of OT regularization over the baseline GFlowNet model only trained with trajectory balance loss on a wide range of tasks: hypergrid environment, discrete probabilistic modeling, and biological sequence design tasks. We aim to show that: (i) Minimizing the path regularization via OT improves the GFlowNets' generalization, and the upper bound can be used as an efficient approximation when we want to minimize the regularization term (hyper-grid environment, discrete probabilistic modeling); while (ii) Maximizing path regularization via OT enhances the exploration ability of GFlowNets (biological sequence tasks). 
\subsection{Hyper-grid environment}
\label{ex: Hyper-grid environment}
\textbf{Task} We follow the framework of \cite{Malkin2022TrajectoryBI} with slight changes to study a hyper-grid environment, which evaluates the generalization ability of the GFlowNet to guess and sample unvisited modes of the interested distribution. Consider a $D$-dimensional hyper-grid environment with length of each side is $H$, where each cell represents non-terminal state of the given DAG: $s = \left(s_{1}, \ldots, s_{D}\right)$ where $s_{d} \in\{0,1, \ldots, H-1\}$ for $d \in \{1, \ldots, D\}$. The source state is $(0,0,...,0)$. For any non-terminal state, the available actions are operations of increasing coordinate $i$ by $1$ that still satisfies $s_{i} \leq H - 1$ and a terminating action that moves to a corresponding terminal state $s^{T}$, which has its reward:
% \begin{equation}
%     \begin{aligned}
%         R\left(s^{\top}\right)=R_{0} &+0.5 \prod_{d=1}^{D} \mathbb{I}\left[\left|\frac{s_{d}}{H-1}-0.5\right| \in(0.25,0.5 ]\right]\\
% &+2 \prod_{d=1}^{D} \mathbb{I}\left[\left|\frac{s_{d}}{H-1}-0.5\right| \in(0.3,0.4)\right]
%     \end{aligned}
% \end{equation}
\begin{equation}
\small
    \begin{aligned}
        R\left(s^{\top}\right)=R_{0} &+0.5 \prod_{d=1}^{D} \mathbb{I}\left[\left|s_{d}/(H-1)-0.5\right| \in(0.25,0.5 ]\right]
+2 \prod_{d=1}^{D} \mathbb{I}\left[\left|s_{d}/(H-1)-0.5\right| \in(0.3,0.4)\right]
    \end{aligned}
\end{equation}
where $R_{0}$ is the constant that controls the discovery challenge and $\mathbb{I}$ is the indicator function. This reward function indicates that only considerable rewards exist at the environment's corners, and there are correct $2^{D}$ modes. The experiment is conducted for two hyper-grid environments with the number of dimensions $4$ and $7$ (a higher number of dimensions means more challenging). We consider the same side length $H=8$ and $R_{0}=10^{-3}$ for both environments. To evaluate the performance on this task, we use KL divergence and the number of modes found during training as the main evaluation metrics. More details about architectures, hyper-parameters, and evaluation criteria are provided in Appendix \ref{appendix:hypergrid}.

\begin{figure}[!ht] 
    \tiny
    \centering
    \includegraphics[width=0.9\linewidth]{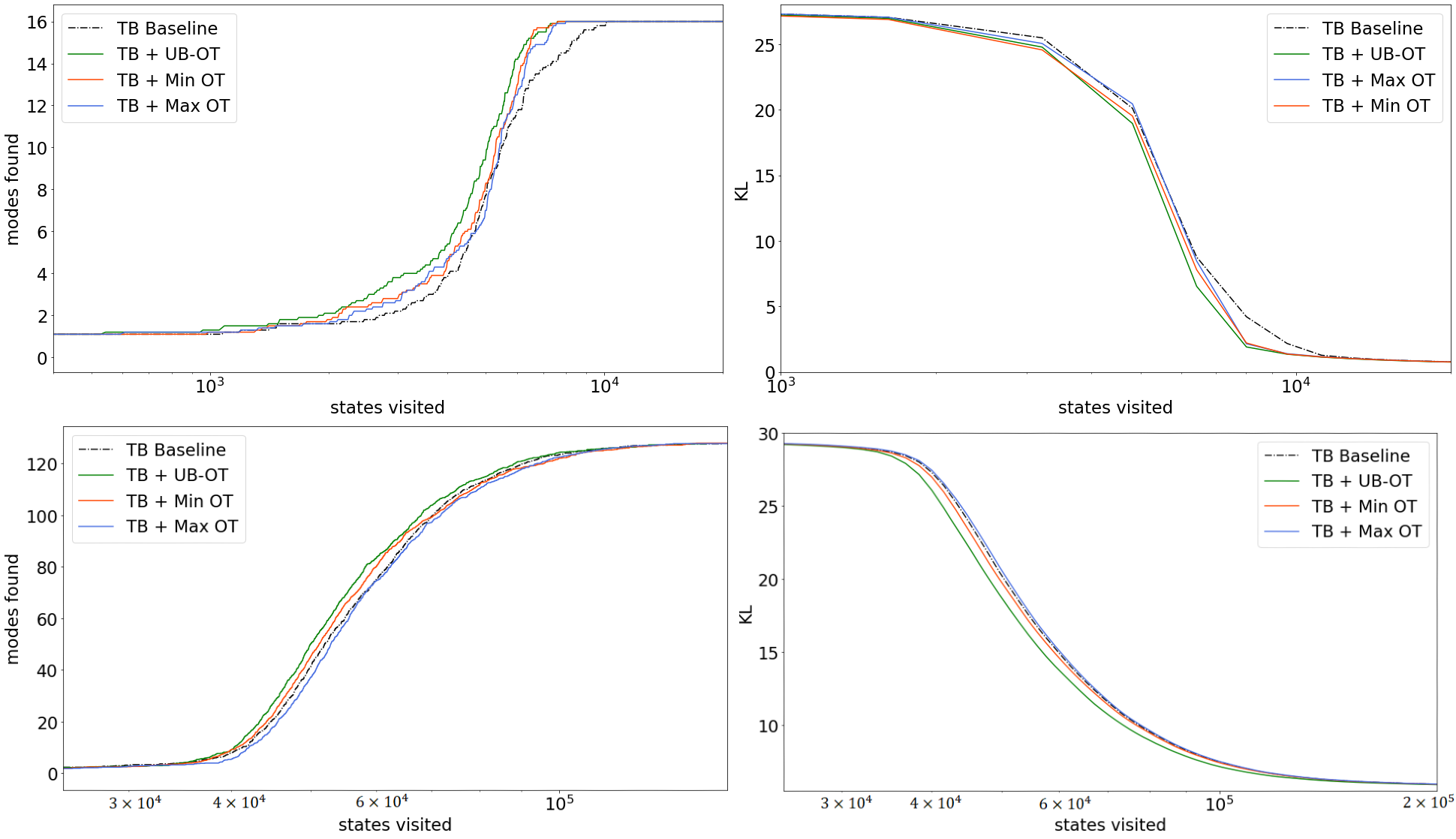}
    \vspace{-0.1in}
    \caption{Results on the $4-D$ (upper) and $7-D$ (lower) hyper-grid environment. Left: Number of modes found during training. Right: KL divergence between the true and empirical distribution.}
    \label {fig:grid}
\end{figure}

\vspace{0.5 em}
\noindent
\textbf{Results} We plot the mean results over $10$ runs for each configuration in Fig. \ref{fig:grid}. %Fig. \ref{fig:grid_4_dim} and  Fig. \ref{fig:grid_7_dim}. 
Although recovering full modes and achieving the same stable minimum of KL divergence, the GFlowNet model trained by minimizing the path regularization via OT discovers modes faster than the baseline, which indicates its focus on finding directions leading to states with high rewards during the training progress rather than spend time exploring the environment. This also helps the model better discover the latent structures of the interested distribution and achieve some level of low KL error faster. We can also see that the upper bound is an efficient approximation in terms of complexity when using a positive regularization coefficient, whose performance is even better. 

% \begin{figure}[!ht] 
%     \centering
%     \includegraphics[width=1\linewidth]{section/figure/experiment/grid_4_dim.png}
%     \caption{Results on the $4-D$ hyper-grid environment. Left: Number of modes found during training. Right: KL divergence between the true and empirical distribution.}
%     \label {fig:grid_4_dim}
% \end{figure}

% \begin{figure}[!ht] 
%     \centering
%     \includegraphics[width=1\linewidth]{section/figure/experiment/grid_7_dim.png}
%     \caption{Results on the $7-D$ hyper-grid environment. Left: Number of modes found during training. Right: KL divergence between the true and empirical distribution.}
%     \label {fig:grid_7_dim}
% \end{figure}

\subsection{Biological sequence design}
\label{exp: Biological sequence design }
\textbf{Task} We follow the framework of \cite{Jain2022BiologicalSD} to simulate the process of designing biological sequences, such as anti-microbial peptides (AMP), DNA and protein sequences (TF Bind 8, GFP). The experiments are conducted in the multi-round active learning setting, with the goal of generating a diverse set of useful candidates after evaluation rounds. We report the performance score, diversity score, and novelty score of the TopK scoring candidates to evaluate the performance of each method. More details about task description, datasets, hyper-parameters, and evaluation criteria are provided in Appendix \ref{appendix:bioseq}.
\begin{table}[!ht]
%\vskip 0.05in
\begin{center}
\tiny
\begin{tabular}{lllll}
\toprule
  & \textbf{Performance} & \textbf{Diversity} & \textbf{Novelty}\\
\midrule
\textbf{DynaPPO} & $\mathbf{0.938 \pm 0.009}$ & $12.12 \pm 1.71$ & $9.31 \pm 0.69$\\
\textbf{COMs} & $0.761 \pm 0.009$ & $19.38 \pm 0.14$ & $26.47 \pm 1.30$\\
\textbf{GFlowNet-AL (paper)} & $0.932 \pm 0.002$ & $22.34 \pm 1.24$ & $28.44 \pm 1.32$\\
\midrule
\textbf{GFlowNet-AL} & $0.874  \pm  0.022$ & $\mathbf{31.98  \pm  2.27}$ &$23.91  \pm  1.87$\\
\textbf{GFlowNet+Min OT-AL} & $0.847 \pm 0.033$ & $20.32  \pm  7.38$ &$23.63  \pm  1.66$\\
\textbf{GFlowNet+UB OT-AL} & $0.828  \pm  0.022$ & $29.89  \pm  2.80$ &$24.16  \pm  1.75$\\
\textbf{GFlowNet+Max OT-AL} & $0.917  \pm  0.003$ & $\mathbf{31.56  \pm  2.43}$ &$\mathbf{28.86  \pm  0.96}$\\
\bottomrule
\end{tabular}
\caption{Results on the AMP task with $K = 100$.}
\label{table:amp_top_100}
\end{center}
%\vskip 0.15in
\vspace{-0.15in}

\end{table}

\vspace{0.5 em}
\noindent
\textbf{AMP} The results for AMP design task is shown in Table \ref{table:amp_top_100}. We see that the GFlowNet-AL model trained by maximizing the regularization via OT performs better than other baselines in terms of diversity and novelty. In addition, the TopK performance of the GFLowNet-AL baseline also increases from $0.874$ to $0.917$ when we maximize the OT regularization and is only lower than the reported performance of DynaPPO. However, DynaPPO has a much lower diversity and novelty score, which implies that it mostly generates similar candidates from the training dataset.

\vspace{0.5 em}
\noindent
\textbf{TF Bind 8} An interesting observation here is that initial dataset $\mathcal{D}_{0}$ contains only half of all possible DNA sequences of length $8$ having lower scores. Specifically, low-quality data is very common in practice, and in this task, it poses a big challenge for all the methods to have good results. From Table \ref{table:tfbind8_top_128}, we can see that MINs have the highest diversity compared to the other methods. However, this method has a much lower TopK performance and novelty score, which indicates its generated samples are very similar to the low-quality training dataset. Moreover, although having slightly lower diversity, the GFlowNet-AL model trained by maximizing the path regularization via OT performs better than the others when looking at all metrics - it outperforms other baselines in terms of performance and novelty score.

\begin{table}[!ht]
\vspace{-0.1in}
\begin{center}
\begin{tabular}{lllll}
\toprule
  & \textbf{Performance} & \textbf{Diversity} & \textbf{Novelty}\\
\midrule
\textbf{DynaPPO} & $0.58 \pm 0.02$ & $5.18 \pm 0.04$ & $0.83 \pm 0.03$\\
\textbf{COMs} & $0.74 \pm 0.04$ & $4.36 \pm 0.24$ & $1.16 \pm 0.11$\\
\textbf{BO-qEI} & $0.44 \pm 0.05$ & $4.78 \pm 0.17$ & $0.62 \pm 0.23$\\
\textbf{CbAS} & $0.45 \pm 0.14$ & $5.35 \pm 0.16$ & $0.46 \pm 0.04$\\
\textbf{MINs} & $0.40 \pm 0.14$ & $\mathbf{5.57 \pm 0.15}$ & $0.36 \pm 0.00$\\
\textbf{CMA-ES} & $0.47 \pm 0.12$ & $4.89 \pm 0.01$ & $0.64 \pm 0.21$\\
\textbf{AmortizedBO} & $0.62 \pm 0.01$ & $4.97 \pm 0.06$ & $1.00 \pm 0.57$\\
\textbf{GFlowNet-AL (paper)} & $0.84 \pm 0.05$ & $4.53 \pm 0.46$ & $2.12 \pm 0.04$\\
\midrule
\textbf{GFlowNet-AL} & $0.83  \pm  0.01$ & $4.66  \pm  0.08$ &$1.14  \pm  0.03$\\
\textbf{GFlowNet+Min OT-AL} & $0.82  \pm  0.01$ & $4.72  \pm  0.10$ &$1.13  \pm  0.04$\\
\textbf{GFlowNet+UB OT-AL} & $0.83  \pm  0.01$ & $4.68  \pm  0.10$ &$1.14  \pm  0.05$\\
\textbf{GFlowNet+Max OT-AL} & $\mathbf{0.85  \pm  0.02}$ & $4.52 \pm 0.18$ &$\mathbf{1.21 \pm 0.10}$\\
\bottomrule
\end{tabular}
\caption{Results on the TF Bind 8 task with $K = 128$.}
\label{table:tfbind8_top_128}
\end{center}

\end{table}

\vspace{0.5 em}
\noindent
\textbf{GFP} Lastly, the results for the GFP task are shown in Table \ref{table:gfp_top_128}, where the objective is to find protein sequences having high fluorescence. We observe that the GFlowNet-AL model trained by maximizing OT regularization generates more diverse and novel candidates than other methods. In addition, its performance score is only lower than the best one achieved by COMs and higher than the GFlowNet-AL baseline. However, when looking at all metrics, the GFlowNet-AL model trained by maximizing the path regularization via OT still outperforms all other baselines.
\begin{table}[!ht]
%\vskip 0.05in
\begin{center}
\begin{tabular}{lllll}
\toprule
  & \textbf{Performance} & \textbf{Diversity} & \textbf{Novelty}\\
\midrule
\textbf{DynaPPO} & $0.794 \pm 0.002$ & $206.19 \pm 0.19$ & $203.20 \pm 0.47$\\
\textbf{COMs} & $\mathbf{0.831 \pm 0.003}$ & $204.14 \pm 0.14$ & $201.64 \pm 0.42$\\
\textbf{BO-qEI} & $0.045 \pm 0.003$ & $139.89 \pm 0.18$ & $203.60 \pm 0.06$\\
\textbf{CbAS} & $0.817 \pm 0.012$ & $5.42 \pm 0.18$ & $1.81 \pm 0.16$\\
\textbf{MINs} & $0.761 \pm 0.007$ & $5.39 \pm 0.00$ & $2.42 \pm 0.00$\\
\textbf{CMA-ES} & $0.063 \pm 0.003$ & $201.43 \pm 0.12$ & $203.82 \pm 0.09$\\
\textbf{AmortizedBO} & $0.051 \pm 0.001$ & $205.32 \pm 0.12$ & $202.34 \pm 0.25$\\
\textbf{GFlowNet-AL (paper)} & $0.853 \pm 0.004$ & $211.51 \pm 0.73$ & $210.56 \pm 0.82$\\
\midrule
\textbf{GFlowNet-AL} & $0.82318  \pm  0.00005$ & $218.54  \pm  7.88$ &$222.05  \pm  5.49$\\
\textbf{GFlowNet+Min OT-AL} & $0.82311  \pm  0.00009$ & $182.03  \pm  0.25$ &$220.56  \pm  2.02$\\
\textbf{GFlowNet+UB OT-AL} & $0.82316  \pm  0.00009$ & $221.64  \pm  0.11$ &$218.02  \pm  0.79$\\
\textbf{GFlowNet+Max OT-AL} & $0.82326  \pm  0.00009$ & $\mathbf{225.00 \pm 3.76}$ &$\mathbf{242.11 \pm 1.44}$\\
\bottomrule
\end{tabular}
\caption{Results on the GFP task with $K = 128$.}
\label{table:gfp_top_128}
\end{center}
%\vskip 0.15in
\end{table}
\subsection{Synthetic discrete probabilistic modeling tasks}
\label{exp: Synthetic discrete probabilistic modeling tasks}
% \cite{pmlr-v162-zhang22v} proposes energy-based generative flow networks (EB-GFN), a probabilistic modeling algorithm for high-dimensional discrete data. They use the Energy-Based Models (EBMs), which capture a distribution over a space $\mathcal{X}$ by associating a scalar energy to each configuration with a density \begin{equation}
%     p_{\phi}(\mathbf{x}) =\frac{1}{Z_{\phi}}\exp(-\mathcal{E}_{\phi}(\mathbf{x}))
% \end{equation}
% Learning the Energy-Based Models requires sampling from the energy distribution. Thus,
% the EB-GFN proposes a  framework to jointly train a GFlowNet with an energy function, so that the GFlowNet learns to sample from the energy distribution, while the energy learns with an approximate MLE objective with negative samples from the GFlowNet. 

% \subsubsection{Task description}
% In this experiment, we follow the framework of \cite{pmlr-v162-zhang22v} to model seven different distributions over 32-dimensional binary data that are discretizations of continuous distributions over the plane. 

% we consider the data is in $D$-dimensional binary space ($ D=32$). The state space $\mathcal{S}$ of the GFlowNet consists of vectors of length $D=32$ with with entries in $\{0; 1;\oslash\}$. The source state is $s_0 = (\oslash,\oslash,...,\oslash)$. For any non-terminal state, the available actions are turning a void entry $\oslash$ to $0$ or $1$. After $D$ actions, we reach the terminal states having all entries in $\{0; 1\}$.

\textbf{Task} We follow the framework of \cite{pmlr-v162-zhang22v}, Energy-based Generative Flow Networks (EB-GFN), to model seven different distributions over 32-dimensional binary data that are discretizations of continuous distributions over the plane. The state space $\mathcal{S}$ of the GFlowNet consists of vectors of length $D=32$ with with entries in $\{0; 1;\oslash\}$. The source state is $s_0 = (\oslash,\oslash,...,\oslash)$. For any non-terminal state, the available actions are turning a void entry $\oslash$ to $0$ or $1$. After $D$ actions, we reach the terminal states having all entries in $\{0; 1\}$. The main evaluation metrics are NLL score and MMD score. The detailed settings about architectures, hyper-parameters, and evaluation criteria are provided in Appendix \ref{appendix:discrete_modelling}.

\vspace{0.5 em}
\noindent
\textbf{Results} The results for synthetic discrete probabilistic modeling tasks (Synthetic EB-GFN) are shown in Table \ref{table:systhetic EB-GFN}. Training GFlowNets with either minimizing the path regularization via OT (Min OT) or via the Upper-bound (UB OT) gains the better NLL and MMD scores than the baseline. We also observe that the performance of training EB-GFN with Min OT and UB OT are quite similar.  
\begin{table}[!ht]
\vskip 0.05in
\begin{center}
\tiny
\begin{tabular}{lllllllll}
\toprule
 \textbf{Metrix} &\textbf{Method}& \textbf{2spirals} & \textbf{8gaussians} & \textbf{circles} &\textbf{moons}& \textbf{pinwheel} & \textbf{swissroll} & \textbf{checkerboard}  \\
\midrule
NLL $\downarrow$ &PCD &$20.094$& $19.991$& $ 20.565$& $ 19.763$& $ 19.593 $& $20.172 $& $21.214$\\
&ALOE &$20.295$&$ 20.350$&$ 20.565$&$ 19.287$&$ 19.821 $&$20.160 $&$54.653$\\
&ALOE + &\bm{$20.062$}&$19.984 $&$20.570$&$ 19.743$&$ 19.576$&$ 20.170$&$ 21.142$\\
&EB-GFN (paper) & $20.050$ & $19.982$  & $20.546$ & $19.732$ & $19.554$  & $20.146$ & $20.696$\\
&EB-GFN & $20.0679$ & $19.9862$  & \bm{$20.5598$} & $19.7324$ & $19.5735$  & $20.1599$ & $20.6839$\\
&EB-GFN + Min OT &$20.0640$ &$19.9855$ &\bm{$20.5598$} &$19.7308$&\bm{$19.5699$}&\bm{$20.1595$}&\bm{$20.6831$}\\
&EB-GFN + UB OT &$20.0651$&\bm{$19.9854$}&$20.5600$&\bm{$19.7305$}&$19.5707$&$20.1596$&$20.6836$\\
\midrule
MMD $\downarrow$ &PCD &$2.160$&$ 0.954 $&$0.188$&$ 0.962$&$ 0.505$&$ 1.382$&$2.831$\\
&ALOE &$21.926$&$ 107.320$&$ 0.497$&$ 26.894$&$ 39.091$&$ 0.471$&$ 61.562$\\
&ALOE + &\bm{$0.149 $}&$0.078$&$ 0.636 $&$0.516 $&$1.746$&$ 0.718 $&$12.138$\\
&EB-GFN (paper) &$0.583$ &$0.531$ &$0.305$ & $0.121$ &$0.492$ &$0.274$ & $1.206$\\
&EB-GFN &$0.3012$ &$0.0408$ &$-0.1724$ & $-0.1744$ &$0.2056$ &$0.1555$ & $-0.0986$\\
&EB-GFN + Min OT &$0.1816$	&$0.0343$ &	\bm{$-0.2775$}&	\bm{$-0.1966$}	&\bm{$0.1220$}&	$0.1334$	&\bm{$-0.1071$}\\
&EB-GFN + UB OT & $0.2902$ &\bm{$0.0102$} &$0.2819$ &$-0.1253$ &$0.1561$ &\bm{$0.0257$}& $-0.0923$\\
\bottomrule
\end{tabular}
\caption{Results on the Synthetic EB-GFN tasks. We display the negative log-likelihood (NLL) and MMD (in units of $1 \times 10^{-4})$. Note that ALOE+ uses a thirty times larger parametrization than ALOE and EB-GFN.}
\label{table:systhetic EB-GFN}
\end{center}
\end{table}

\section{Concluding Remarks}
\label{sec:conclusion}
In this paper, we propose to train GFlowNets with an additional path regularization via Optimal Transport that places prior constraints on the underlying structure of the GFlowNets. We have empirically shown that minimizing the path regularization via OT improves the GFlowNet’s generalization while maximizing path regularization via
OT enhances the exploration ability of GFlowNets. In addition, we derive an efficient implementation of the regularization by finding its closed form solutions in specific cases and a meaningful upper bound that can be used as an approximation when we want to
minimize the regularization term. A limitation of the current method is computing the optimal transport distances for all couples of nearest neighbor states. Our proposed Dropout OT (see in Appendix \ref{Dropout Optimal Transport}) might be a solution. In future works, we aim to develop a more efficient path regularization for high dimensional discrete data, which potentially can be addressed by using sliced optimal transport~\cite{bonneel2015sliced, nguyen2022revisiting, nguyen2022_hierarchical}, or propose a new cost function to compute the optimal transport distances.

\bibliography{iclr2022_conference}
\bibliographystyle{abbrv}

%Transport plan as Joint distribution

%Optimal transport problem

%Skrinkorn

\clearpage
\appendix
\begin{center}
{\bf{\Large{Supplement to ``Improving Generative Flow Networks with Path Regularization"}}}
\end{center}

\section{Related Work}
\label{sec:appendix_related_work}

\textbf{GFlowNets} The objective of GFlowNets is related to MCMC methods for sampling from a given unnormalized density function, especially in discrete spaces where exact sampling is intractable (\cite{Dai2020LearningDE, Grathwohl2021OopsIT}). However, GFlowNets amortize the complexity of iterative sampling by a training procedure that implies the data's compositional structure as its learning problem. Empirically, GFlowNets' performance is better than other earlier methods in a wide variety of tasks: small molecules generation (\cite{DBLP:journals/corr/abs-2106-04399}), discrete probabilistic modeling (\cite{pmlr-v162-zhang22v}), Bayesian structure learning (\cite{Deleu2022BayesianSL}) and biological sequence design (\cite{Jain2022BiologicalSD}). On the theoretical side, definitions and properties of GFlowNets are more investigated in \cite{https://doi.org/10.48550/arxiv.2111.09266}.

\vspace{0.5 em}
\noindent
\textbf{Optimal Transport} The optimal transport theory (OT) (\cite{Villani2003TopicsIO}) has established a natural and useful geometric tool for comparing measures supported on metric probability spaces. The development of OT theory has a long history, where it has been discovered in many settings and under different forms. And in recent years, another revolution in the spread of OT has been witnessed, thanks to the emergence of approximate solvers that can scale to the problem of large dimensions. As a consequence, OT is being widely used to solve various problems in computer graphics (\cite{Bonneel2011DisplacementIU},\cite{Nguyen_3D}), image processing (\cite{Xia2014SynthesizingAM}), and machine learning (\cite{Courty2014DomainAW}, \cite{ho2017multilevel} \cite{Genevay2018LearningGM}, \cite{Bunne2019LearningGM}).

\vspace{0.5 em}
\noindent
\textbf{Energy-based models} EBMs, or energy functions parameterized by deep neural networks, have demonstrated effectiveness in generative modeling (\cite{Salakhutdinov2009DeepBM, Hinton2006AFL}). Contrastive divergence methods (\cite{Hinton2002TrainingPO, Tieleman2008TrainingRB, Du2021ImprovedCD}) have been proposed to handle costly MCMC processes by approximating energy gradient. Recently, it has been shown that simultaneous learning of the proposal distribution can also be helpful (\cite{Dai2019ExponentialFE, Arbel2021GeneralizedEB}). Then this finding has been extended to discrete spaces by using GFlowNets in \cite{pmlr-v162-zhang22v}.

\vspace{0.5 em}
\noindent
\textbf{Biological sequence design} Various methods have been proposed to handle the biological sequence design tasks: deep model-based optimization (\cite{Trabucco2021ConservativeOM}), Bayesian optimization (\cite{Belanger2019BiologicalSD, Swersky2020AmortizedBO}), reinforcement learning (\cite{Angermueller2020ModelbasedRL}), adaptive evolutionary methods (\cite{Hansen2006TheCE, Sinai2020AdaLeadAS}), and so on. Recently, GFlowNets also have been proposed as a useful generator of diverse candidates for this problem in \cite{Jain2022BiologicalSD}. 

\section{Background of GFlowNets}
\label{sec:appendix_background}

Generative Flow Networks (GFlowNets) are a class of generative model learning to sample a compositional object $\mathbf{x}$ with probability in proportion to a given reward function $R(\mathbf{x})$.
From a reinforcement learning viewpoint, GFlowNets learn a stochastic policy to generate object $\mathbf{x}\in \mathcal{X}$ by applying a sequence of discrete actions $a\in \mathcal{A}$. After an action, we might have an object or partially constructed object, which is described by a state $s\in \mathcal{S}$. The sequence of actions forms a complete trajectory $\tau = (s_0, s_1,...,s_n,s_f)$ specifies how the object was build from scratch. The trajectory starting from an initial empty state $s_0$, going through a sequence of partially constructed object $s$ and terminating at constructed object $s_n = x$ to a special final state $s_f$. The fully constructed objects in $\mathcal{X} \subset \mathcal{S}$ are called terminating states. From a probabilistic modeling viewpoint, GFlowNets can be seen as the flows of unnormalized probabilities in a directed acyclic graph $G = (\mathcal{S},\mathcal{A})$ from a source $s_0$ to sinks $\mathbf{x} \in \mathcal{X}$, which then flows to the final single sink state $s_f$. The stochastic policy in GFlowNets acts as the gates to control the flows of unnormalized probabilities. If GFlowNets modify the gates so that the flow of unnormalized probabilities
leading to the sink $\mathbf{x}$ is equals to the given reward $R(\mathbf{x})$, they can sample object $\mathbf{x}$ with probability in proportion to $R(\mathbf{x})$.

A \textit{complete trajectory} is a sequence of states $\tau = (s_0, s_1,...,s_n,s_f)$, where each transition $s_{t}\to s_{t+1}$ is an action in $\mathcal{A}$. For each transition $s\to s'\in \mathcal{A}$, we call $s$ is a parent of $s'$ and $s'$ is  a child of $s$. Let $\mathcal{T}$ be the set of complete trajectories.

\vspace{0.5 em}
\noindent
\textbf{Flows} A \textit{trajectory flow}~\cite{https://doi.org/10.48550/arxiv.2111.09266} is a nonnegative function $F: \mathcal{T} \mapsto \mathbb{R^+}$.
The flow $F(\tau )$ can be interpreted as the probability mass associated with trajectory $\tau$. As an analogy with water flow, the flow $F(\tau )$ is the number of water particles sharing the same path $\tau$. 

The flow through a state (state flow) is defined as follows:
\begin{equation}
    F(s) = \sum_{s\in \tau}F(\tau).
\end{equation}

The flow through a edge (edge flow) is defined as follows:
\begin{equation}
    F(s\rightarrow s') = \sum_{ s \rightarrow s' \in \tau}F(\tau).
\end{equation}
Trajectory flow is a measure on the set of all complete trajectories $\mathcal{T}$. Thus we can associate a probability
measure $P$ with the trajectory flow $F$:
\begin{equation}
    P(\tau) :=\frac{F(\tau)}{\sum_{\tau \in \mathcal{T}}F(\tau)} =  \frac{F(\tau)}{Z}.
\end{equation}
We can also define the probability $P_T(x)$ of visiting a terminal state $x$ as follows:
\begin{equation}
    P_T(x) :=\frac{\sum_{\tau: x \in \tau }F(\tau)}{Z} = \frac{F(x)}{Z}. %=\sum_{\tau: x \in \tau }{P(\tau)}
\end{equation}
The probability of an event $A\subseteq \mathcal{T}$ is defined as follows:
\begin{equation}
       P(A) := \frac{F(A)}{Z}
\end{equation}
where $F(A) =\sum_{\tau \in A}{F(\tau)}$.

The conditional probability of an event A given another event B is defined as follows:
\begin{equation}
    P(A|B) := \frac{P(A, B)}{P(B)} = \frac{F(A\cap B)}{F(B)}.
\end{equation}
The forward transition probabilities (forward policy) $P_F$ and the backward transition probabilities (backward policy) are special cases of conditional probability, which are defined as follows:
\begin{equation}\label{Forward}
    P_{F}(s'|s) :=P(s\to s'|s) = \frac{F(s\to s')}{F(s)}, 
\end{equation}
\begin{equation}
    P_{B}(s|s') :=P(s\to s'|s') = \frac{F(s\to s')}{F(s')}.
\end{equation}

\vspace{0.5 em}
\noindent
\textbf{Markovian Flow} Trajectory flow $F$ is Markovian Flow if, for any state $s$, outgoing edge $s\to s'$, and for any
trajectory $\tau = (s_1, \dots , s)$ ending in s:
\begin{equation}
    P(s\to s'|\tau) = P(s\to s'|s) = P_F(s'|s).
\end{equation}
Equivalently, from Proposition 3 in~\cite{DBLP:journals/corr/abs-2106-04399}, we have
\begin{equation}\label{GFlowNets genarate}
    P(\tau = (s_{0} \rightarrow ... \rightarrow s_{n})) = \prod_{t=1}^{n}P_{F}(s_{t}\mid s_{t-1}),
\end{equation}
or equivalently
\begin{equation}
    P(\tau = (s_{0} \rightarrow ... \rightarrow s_{n} \mid s_{n}=x)) = \prod_{t=1}^{n}P_{B}(s_{t-1}\mid s_{t}).
\end{equation}
Because each trajectory in GFlowNets was generate by policy $\pi(s'|s) = P_F(s'|s)$ via Eqn.~\eqref{GFlowNets genarate}, as in Corollary 2~\cite{DBLP:journals/corr/abs-2106-04399}, if a GFlowNet yields a flow, then it is a Markovian flow. Learning a GFlowNet is considered the problem of fitting a Markovian flow to a fixed reward function on $\mathcal{X}$ as below.

\vspace{0.5 em}
\noindent
\textbf{Learning GFlowNets~\cite{DBLP:journals/corr/abs-2106-04399}} Suppose that a nontrivial nonnegative reward function $R: \mathcal{X} \rightarrow \mathbb{R}_{\geq0}$ is given on the set of terminal states. GFlowNets aim to approximate a Markovian flow $F$ on $G = (\mathcal{S},\mathcal{A})$ such that the likelihood of
a trajectory sampled from $F$ terminating in a given $x \in
\mathcal{X}$ is proportional to $R(x)$, i.e. $P_T(x) \propto R(x) $. This is
accomplished by the reward matching condition:
\begin{equation}\label{flow-matching}
    R(x) = \sum_{\tau: x\in \tau}{F(\tau)} = F(x) \quad \forall x \in \mathcal{X}.
\end{equation}

If $R(x) = \sum_{\tau: x\in \tau}{F(\tau)} = F(x)$ then $P_T(x) = \frac{F(x)}{Z} \propto R(x) $, i.e. the likelihood of a trajectory sampled from $F$ terminating in a given $x \in
\mathcal{X}$ is proportional to $R(x)$.
GFlowNet has several training objectives which were proven to satisfy Eqn.~\eqref{flow-matching}.

\vspace{0.5 em}
\noindent
\textbf{Flow matching objective~\cite{DBLP:journals/corr/abs-2106-04399}:}
\begin{equation}\label{Flow matching condition}
\sum_{s'}F(s'{\rightarrow}s)=\sum_{s''} F(s{\rightarrow}s''). 
\end{equation}
This is the conservation law, i.e., the total flow into a state is equal to the total flow moving out of it. We can parameterize GFlowNets with the edge flow function $F_{\theta}(\cdot,\cdot)$, and sample object $x$ via the policy as the forward transition probabilities defined by this flow:
\begin{equation}
    \pi(s'|s) = P_{F_{\theta}}(s'|s)=\frac{F_{\theta}(s{\rightarrow}s')}{\sum_{s''} F_{\theta}(s{\rightarrow}s'')}.
\end{equation}

~\cite{DBLP:journals/corr/abs-2106-04399} shows that if the flow matching objective obtains a global minimum over a full-support training distribution $\overline{\pi}_{\theta}$, the policy $\pi(s'|s)$ would sample each
object $x$ with a probability $P_T(x)$ proportional to its reward R(x).
\begin{equation}
\mathcal{L}_{FM}=\left(\log(\sum_{s'} F_{\theta}(s'{\rightarrow}s))-\log(\sum_{s''}F_{\theta}(s{\rightarrow}s''))\right)^2.
\end{equation}

\vspace{0.5 em}
\noindent
\textbf{Detailed balance objective}:
\begin{equation}F(s)P_F(s'|s) = P_B(s|s') F(s').\end{equation}
This is because both sides of the term equal $F(s\to s')$.
The detailed balance condition has a close relation with the flow matching condition, where the sum of flow is implicitly included in $P_F$ and $P_B$. We can parameterize
GFlowNets with 
$F_{\theta}(s)$, $P_{F}(\cdot|s;\theta)$, and $P_{B}(\cdot|s';\theta)$.
As flow matching condition, we can turn detailed balance condition into a training objective:
\begin{equation}
\mathcal{L}_{DB}=\left(\log( F_{\theta}(s)P_{F}(\cdot|s;\theta))-\log(F_{\theta}(s')P_{B}(\cdot|s';\theta))\right)^2.
\end{equation}

\vspace{0.5 em}
\noindent
\textbf{Trajectory balance objective}
For any trajectory $\tau = (s_0,s_1,...s_n=x)$
\begin{equation}
F(s_0)\prod_{t=1}^n P_F(s_t|s_{t-1}) = R(x) \prod_{t=1}^n P_B(s_{t-1}|s_t).\end{equation}
This is because both sides of the term equal $F(\tau)$. We can parameterize
GFlowNets with 
$Z_\theta = F(s_0)$, $P_{F}(\cdot|s;\theta)$, and $P_{B}(\cdot|s';\theta)$.
As flow matching condition, we can turn trajectory balance condition into a training objective:
\begin{equation}
\mathcal{L}_{TB}=\left(\log(Z_\theta\prod_{t=1}^n P_F(s_t|s_{t-1};\theta) )-\log(R(x) \prod_{t=1}^n P_B(s_{t-1}|s_t;\theta))\right)^2.
\end{equation}
Trajectory balance objective brings more efficient credit assignment and faster convergence~\cite{Malkin2022TrajectoryBI}.

\vspace{0.5 em}
\noindent
\textbf{GFlowNets can be trained offline}
This means we can use any train policy $\overline{\pi}$ of sampling trajectories $\tau$ to minimize the training objective $\mathbb{E}({\mathcal{L}(\tau)})$ so long as it can sample all the possible trajectories with non-zero probability.

Usually, training policy $\overline{\pi}$ was chosen as a mixing of GFLowNets policy $\pi(s'|s)$ with $\alpha$ percentage of exploration by a uniform distribution:
\begin{equation}
    \overline{\pi} = (1-\alpha)\pi(s'|s)+\alpha\text{Uniform}.
\end{equation}

\section{Dropout Optimal Transport}
\label{Dropout Optimal Transport}
A limitation of the current method is computing the optimal transport distances for all couples of nearest neighbor states, especially in high dimensional discrete data. Our proposed Dropout OT might be a solution. This is because rather than sampling trajectories $\tau$ and using all edges from them, we can separately sample edges $s\to s'$ proportional to edge flows, allowing us to efficiently compute path regularization.

\begin{theorem}\label{theorem: dropout OT}
For any complete trajectory $\tau = (s_{0}\rightarrow s_{1} \rightarrow ... \rightarrow s_{n})$ sampled from the training policy $\pi_{\theta}$
\begin{equation}
    \mathbb{E}_{\tau \sim \pi _{\theta}}( \mathcal{L}_{\text{OT}}(\tau)) \propto  \mathbb{E}_{s\to s' \sim  \pi _{\theta}}( \text{OT}\left ( P_F(\cdot| s),P_F(\cdot| s') \right )).
\end{equation}
\end{theorem}
The proof of Theorem \ref{theorem: dropout OT} is in Appendix \ref{proof:dropout OT }.
Here we train GFlowNets with Trajectory Balance Objective. Therefore, when sampling a trajectory $\tau$, we get a set of edges from $\tau$. We just sample uniformly a $p$ percentage of edges to compute OT loss.

To sample $p$ percentage of edges, let sample $r_s \sim Ber(p)$.
\begin{equation}
    \mathbb{E}_{s\to s' \sim  \pi _{\theta}}( \text{OT}\left ( P_F(\cdot| s),P_F(\cdot| s') \right )) = \frac{1}{p}\mathbb{E}_{r_s\sim Ber(p)}\mathbb{E}_{s\to s' \sim  \pi _{\theta}}( r_s.\text{OT}\left ( P_F(\cdot| s),P_F(\cdot| s') \right )).
\end{equation}
We approximate the path regularization loss via:
\begin{equation}
        \mathcal{L}_{\text{OT}}(\tau ) \simeq \frac{1 }{p}\sum_{t=0}^{n-1}x_{t}\text{OT}(P_{F}(.| s_{t}), P_{F}(.|s_{t+1}))
\end{equation}
with $x_{t}$ drawn independently from $\text{Ber}(p)$ for all $0 \leq t \leq n-1$ . Intuitively, if $x_{t}=0$ then we don't need to calculate the corresponding optimal transport cost anymore, which reduces a considerable amount of computing time and memory down to $p$ percentage.

\section{Proofs}
\label{sec:appendix}

\subsection{
Proof of Theorem \ref{thoerem: Upperbound}}\label{proof: upperbound}
For any trajectory $\tau = (s_{0}\rightarrow s_{1} \rightarrow ... \rightarrow s_{n})$, we first prove that for any $t\in \overline{0,n-1}$
\begin{equation}\label{eq: apendix upperbound}
     \text{OT}\left ( P_F(\cdot| s_t),P_F(\cdot| s_{t+1}) \right ) \leq \mathbf{H}(P_{F}(\cdot| s_{t}),P^{*}_{B}(\cdot| s_{t})) - \log(P_{F}(s_{t+1}|s_{t})) + \mathbf{H}(P_{F}(\cdot| s_{t+1})).
\end{equation}

Consider two neigboor states $s_t$ and $s_{t+1}$ with the children sets: $\text{Child}(s_{t})=\{u_{1},...,u_{k}\}$ and $\text{Child}(s_{t+1})=\{v_{1},...,v_{l}\}$.
By Definition \ref{eq: OT distance}, the optimal transportation distance between two distributions $P_{F}(.| s_{t})$ and $P_{F}(.| s_{t+1})$ is defined as:
\begin{equation}
\label{eq: ot cost original in appendix}
    \text{OT}_\mathbf{C}\left ( P_F(\cdot| s_t),P_F(\cdot| s_{t+1}) \right ) :=\min _{\pi \in \prod\left(P_F(\cdot| s_t),P_F(\cdot| s_{t+1})\right)}\langle \mathbf{C},\pi\rangle,
\end{equation}
where the admissible couplings set is defined as:
\begin{equation} \label{eq: constraint coupling}
    \begin{aligned}
        \Pi(P_{F}(.|s_{t}), P_{F}(.|s_{t+1})) = \left \{ \pi \in \mathbb{R}_{+}^{k\times l}: \pi\mathbbm{1}_{l} = P_{F}(\cdot|s_{t}), \pi^{\text{T}}\mathbbm{1}_{k} = P_{F}(\cdot|s_{t+1})\right \}.
    \end{aligned}
\end{equation}

We have,
\begin{equation}
    \begin{aligned}
        &\quad \text{OT}(P_{F}(.|s_{t}), P_{F}(.|s_{t+1}))\\ 
        &\leq \sum _{i}\sum _{j} \pi_{ij}\mathbf{C}_{ij} \\
        & \leq -\sum _{i}\sum _{j} \pi_{ij} \log\left (P_{B}(s_{t} | u_{i})P_{F}(s_{t+1} | s_{t})P_{F}(v_{j} | s_{t+1}) \right )\\
        &= -\sum _{i}\sum _{j} \pi_{ij} \log\left (P_{B}(s_{t}| u_{i})\right ) 
        -\sum _{i}\sum _{j} \pi_{ij} \log\left (P_{F}(s_{t+1} | s_{t})\right )
         -\sum _{i}\sum _{j} \pi_{ij} \log\left (P_{F}(v_{j} | s_{t+1}))\right ) \\
        &= -\sum _{i}\log\left (P_{B}(s_{t}| u_{i})\right )\sum _{j} \pi_{ij} 
        -\log\left (P_{F}(s_{t+1} | s_{t})\right )\sum _{i}\sum _{j} \pi_{ij} 
        -\sum _{i}\log\left (P_{F}(v_{j} | s_{t+1}))\right )\sum _{j} \pi_{ij}  \\
        &= -\sum _{i}\log\left (P_{B}(s_{t}| u_{i})\right )P_{F}(u_{i}| s_{t}) - \log\left (P_{F}(s_{t+1} | s_{t})\right )
        - \sum _{j} \log\left (P_{F}(v_{j} | s_{t+1}))\right )P_{F}(v_{j} | s_{t+1})  \\
        &= -\sum _{i}\log\left (P_{B}^{*}(u_{i}| s_{t})\right )P_{F}(u_{i}| s_{t})  - \log\left (P_{F}(s_{t+1} | s_{t})\right ) + \mathbf{H}(P_{F}(.| s_{t+1}) \\
        &= \mathbf{H}(P_{F}(.| s_{t}),P^{*}_{B}(.| s_{t})) - \log(P_{F}(s_{t+1}|s_{t})) + \mathbf{H}(P_{F}(.| s_{t+1}).
    \end{aligned}
\end{equation}

The first inequality obtained by the definition of optimal transport distance in Eqn.~\eqref{eq: ot cost original in appendix}, the second inequality comes from Eqn.~\eqref{eq: transport cost}, the fifth equality is due to the constraints of admissible couplings in Eqn.~\eqref{eq: constraint coupling}.

As a consequence, the upper-bound loss is obtained by summing up all inequalities~\eqref{eq: apendix upperbound} for all $t$.

\subsection{
Proof of Theorem \ref{theorem: close form}}\label{proof: close form}
Recall from Definition \ref{eq: OT distance}  the optimal transportation distance between two distributions $P_{F}(.| s)$ and $P_{F}(.| s')$ is defined as:
\begin{equation}
\label{eq: ot cost original in appendix 2}
    \text{OT}_\mathbf{C}\left ( P_F(\cdot| s),P_F(\cdot| s') \right ) :=\min _{\pi \in \Pi\left(P_F(\cdot| s),P_F(\cdot| s')\right)}\langle \mathbf{C},\pi\rangle.
\end{equation}
Let decompose the total cost $\langle \mathbf{C},\pi\rangle$ as follows:
\begin{equation}\label{eq: apendix decompose}
    \begin{aligned}
    \langle \mathbf{C},\pi\rangle &= \sum_{i,j} \pi_{ij}\mathbf{C}_{ij} \\&= \sum_{i,j} \pi_{ij}(-\log( P_B(s|u_i))-\log(P_F(s'|s))-\log(P_F(v_j|s'))) \\&+ \sum_{u_i = s', j} \pi_{ij}(\log( P_B(s|s'))+\log(P_F(s'|s))) \\&+\sum_{u_i \neq s', v_j \in Child(u_i), a_i\neq a^\top} \pi_{ij}(\log( P_B(s|u_i))+\log(P_F(s'|s))+\log(P_F(v_j|s')+\mathbf{C}_{ij})
    \\&+\sum_{u_i = v_j} \pi_{ij}(\log( P_B(s|u_i))+\log(P_F(s'|s))+\log(P_F(v_j|s'))).
    \end{aligned}
\end{equation}
We will prove that $u_i \neq v_j\quad \forall i,j$, i.e,  $\text{Child}(s)\cap \text{Child}(s') = \O$, indeed,
\begin{equation}
    a_i \neq a_k+a_h \quad \forall a_i,a_k,a_h \in \mathcal{A} \Longrightarrow a_i \neq a^*_s+a_j\Longrightarrow s+a_i \neq s+a^*_s+a_j \Longrightarrow u_i \neq v_j. \quad \forall i,j
\end{equation}

%Besides, we will prove that $u_i \neq s', v_j \in Child(u_i), a_i\neq a^\top \Longleftrightarrow a_i=a_j\neq a^\top,u_i +a^*_s = v_i $ 
We have
\begin{equation}
\begin{aligned}
    &u_i \neq s',\quad v_j \in Child(u_i),\quad a_i\neq a^\top\\ \Longleftrightarrow \quad &a_i\neq a^*_s,\quad s+a_i+a^*_{u_i}=s+a^*_s+a_j,\quad a_i\neq a^\top
    \\ \Longleftrightarrow \quad &a_i\neq a^*_s,\quad a_i+a^*_{u_i}=a^*_s+a_j,\quad a_i\neq a^\top
    \\ \Longleftrightarrow \quad &a_i\neq a^*_s,\quad a_i=a_j\neq a^\top,\quad a^*_{u_i}=a^*_s.
\end{aligned}
\end{equation}
As a result, we can rewrite Eqn.~\eqref{eq: apendix decompose} as:
\begin{equation}
    \begin{aligned}
    \langle \mathbf{C},\pi\rangle &= \sum_{i,j} \pi_{ij}(-\log( P_B(s|u_i))-\log(P_F(s'|s))-\log(P_F(v_j|s'))) \\&+ \sum_{u_i = s', j} \pi_{ij}(\log( P_B(s|s'))+\log(P_F(s'|s))) \\&+\sum_{u_i \neq s', a_i = a_j\neq a^\top} \pi_{ij}(\log( P_B(s|u_i))+\log(P_F(s'|s))+\log(P_F(v_j|s')+\mathbf{C}_{ii}).
    \end{aligned}
\end{equation}
The first term of above equation actually is the upper-bound of the optimal transport distance. Therefore, we can rewrite the total transport cost as:
\begin{equation}\label{eq: apendix decompose 1}
    \begin{aligned}
    \langle \mathbf{C},\pi\rangle 
    &=\mathbf{H}(P_{F}(\cdot| s),P^{*}_{B}(\cdot|s)) - \log(P_{F}(s'|s)) + \mathbf{H}(P_{F}(\cdot|s')\\
    &+ P_F(s'|s).(\log( P_B(s'|s))+\log(P_F(s'|s)))\\
    &+\sum_{u_i \neq s', a_i = a_j\neq a^\top} \pi_{ij}(\log( P_B(s|u_i))+\log(P_F(s'|s))+\log(P_F(v_j|s')+\mathbf{C}_{ii}).
    \end{aligned}
\end{equation}
From the definition of $c'_i$ in Eqn.~\eqref{eq: def cprime}, we have
\begin{align}\label{eq: apendix decompose 2}
c'_{i}= 
\begin{cases}
  \log( P_B(s|u_i))+\log(P_F(s'|s))+\log(P_F(v_j|s')+\mathbf{C}_{ii}, & \text{if $u_i \neq s', a_i = a_j\neq a^\top$} \\
  0 & \text{if $u_i = s'$ or $a_i = a^\top$}.
\end{cases}
\end{align}
From Eqn.~\eqref{eq: apendix decompose 1} and Eqn.~\eqref{eq: apendix decompose 2}, we have
\begin{equation}
    \sum_{u_i \neq s', a_i = a_j\neq a^\top} \pi_{ij}(\log( P_B(s|u_i))+\log(P_F(s'|s))+\log(P_F(v_j|s')+\mathbf{C}_{ii}) = \sum_{i}{\pi_{ij}.c'_i}.
\end{equation}
Thus, we find that
\begin{equation}
    \mathop{\arg \min}\limits_{\pi \in \prod\left(P_F(\cdot| s),P_F(\cdot| s')\right)}\langle \mathbf{C},\pi\rangle = \mathop{\arg \min}\limits_{\pi \in \prod\left(P_F(\cdot| s),P_F(\cdot| s')\right)}\langle \mathbf{C'},\pi\rangle
\end{equation}
where, $\mathbf{C'}$ is a diagonal matrix with the diagonal $c'_i\leq 0$.  For convenience, if action $a_i$ is invalid at state $s$, we assign $P_F(u_{i}\mid s) := 0$, so the cost matrix of the optimal transport distance still is a square matrix with the zero cost a invalid actions, then applying the Lemma 1, we have: 
\begin{equation}
    \min _{\pi \in \prod\left(P_F(\cdot| s),P_F(\cdot| s')\right)}\langle \mathbf{C'},\pi\rangle = \sum_{i}{\min(P_F(u_i| s),P_F(v_i| s))\mathbf{C}^{'}_{ii}}.
\end{equation}
We obtain the closed form formulation for optimal transport distance
\begin{equation}
\begin{aligned}
    \text{OT} \left ( P_F(\cdot| s),P_F(\cdot| s') \right ) &= \mathbf{H}(P_{F}(\cdot| s),P^{*}_{B}(\cdot| s)) - \log(P_{F}(s'|s)) + \mathbf{H}(P_{F}(\cdot| s'))\\&+P_F(s'|s).(\log( P_B(s'|s))+\log(P_F(s'|s)))
    \\&+\sum_{i\in A^{*}_s\bigcap A^{*}_{s'} }{\min(P_F(u_i|s),P_F(v_i|s'))c'_{i}}.
\end{aligned}
\end{equation}

\begin{lemma}
\label{lemma:key}
Given a squared diagonal cost matrix $\mathbf{C'}$ with non-positive entities in the diagonal, the solution of optimal transport problem between two distribution $P_F(\cdot|s)$ and $P_F(\cdot|s')$, which has the same number of support points, given cost matrix $\mathbf{C'}$ is given by:
\begin{equation}
    \min _{\pi \in \Pi\left(P_F(\cdot| s),P_F(\cdot| s')\right)}\langle \mathbf{C'},\pi\rangle = \sum_{i}{\min(P_F(u_i| s),P_F(v_i| s))\mathbf{C}^{'}_{ii}}.
\end{equation}
\end{lemma}
\noindent
\textbf{Proof of Lemma~\ref{lemma:key}:}
Let define $$F(\pi) = \langle \mathbf{C'},\pi\rangle, $$
\[
\overline{p}_{ij}= 
\begin{cases}
  \min(p_{s}^i,p_{s'}^i), & \text{if $i=j$} \\
  \frac{\left(p_{s}^i-\min(p_{s}^i,p_{s'}^i)\right)\left(p_{s'}^j-\min(p_{s}^j,p_{s'}^j)\right)}{1-\sum_{k}{min(p_{s}^k,p_{s'}^k)}} & \text{if $i\neq j$}.
\end{cases}
\]
where $p_{s}^i :=P_F(u_i| s) $ and $p_{s'}^j :=P_F(v_j| s') $.

We will prove that $\overline{\pi} \in \Pi\left(P_F(\cdot| s),P_F(\cdot| s')\right) $ and $F(\pi)\geq F(\overline{\pi}) \quad \forall \pi \in \Pi\left(P_F(\cdot| s),P_F(\cdot| s')\right) $.

It is not difficult to show that $\overline{\pi}_{ij}\geq 0 $. From the definition of $\overline{\pi}$, we have
\begin{equation}
    \sum_j^n{\overline{\pi}_{ij}} = \sum_{j\neq i}{\overline{\pi}_{ij}}+\overline{\pi}_{ii} = \sum_{j\neq i}{\frac{\left(p_{s}^i-\min(p_{s}^i,p_{s'}^i)\right)\left(p_{s'}^j-\min(p_{s}^j,p_{s'}^j)\right)}{1-\sum_{k}{min(p_{s}^k,p_{s'}^k)}}}+\min(p_{s}^i,p_{s'}^i).
\end{equation}

%For all $i$, two possibilities exist: either $\min(p_{s}^i,p_{s'}^i)=p_{s}^i$ or $\min(p_{s}^i,p_{s'}^i) = p_{s'}^i$

If $\min(p_{s}^i,p_{s'}^i)=p_{s}^i$ then
\begin{equation}
    \sum_j^n{\overline{\pi}_{ij}} = 0+\min(p_{s}^i,p_{s'}^i) = p_{s}^i.
\end{equation}
else $\min(p_{s}^i,p_{s'}^i) = p_{s'}^i$ then
\begin{equation}
    \begin{aligned}
        &\sum_{j\neq i}{\left(p_{s'}^j-\min(p_{s}^j,p_{s'}^j)\right)} = \sum_{j}{\left(p_{s'}^j-\min(p_{s}^j,p_{s'}^j)\right)} = 1-\sum_{k}{\min(p_{s}^k,p_{s'}^k)} \\
        \Longrightarrow &\sum_j^n{\overline{\pi}_{ij}} = \left(p_{s}^i-\min(p_{s}^i,p_{s'}^i)\right){\frac{\sum_{j\neq i}\left(p_{s'}^j-\min(p_{s}^j,p_{s'}^j)\right)}{1-\sum_{k}{\min(p_{s}^k,p_{s'}^k)}}}+\min(p_{s}^i,p_{s'}^i) = p_{s}^i.
    \end{aligned}
\end{equation}

Therefore $\sum_j^n{\overline{\pi}_{ij}} = p_{s}^i = P_F(u_i|s)$. Similarly, $\sum_i^n{\overline{\pi}_{ij}} = p_{s}^j = P_F(v_j|s')$, combining with $\overline{\pi}_{ij}\geq 0$, we have
\begin{equation}
    \overline{\pi} \in \Pi\left(P_F(\cdot| s),P_F(\cdot| s')\right).
\end{equation}

Moreover
\begin{equation}
    F(\pi) = \langle \mathbf{C'},\pi\rangle= \sum_{i} \pi_{ii}\mathbf{C'_{ii}} \geq \sum_{i}{\min(p_{s}^i,p_{s'}^i)\mathbf{C'_{ii}}} = \langle \mathbf{C'},\overline{\pi}\rangle =  F(\overline{\pi}) \quad \forall \pi \in \Pi\left(P_F(\cdot| s),P_F(\cdot| s')\right).
\end{equation}
As a consequence, we obtained the solution of optimal transport problem.

\vspace{1 em}
\noindent
\textbf{Closed form solution for optimal transport distance at terminal state.}
We will derive the closed form solution for optimal transport distance in case of two neighbor states $s < s^{\prime}$, in which $s'$ is a terminal state. In the case of Hyper-grid environment, EB-GFN experiments, and Biological Sequence Design, all terminal state $x$ have only one child that is the final state $s_f$, and $P_F(s_f|x) = 1\quad \forall x$. Thus, the admissible couplings set $\prod(P_{F}(\cdot|s), P_{F}(\cdot|s'))$ has only one element. That is $\pi^* =P_{F}(\cdot|s)$. As a result, the optimal transportation distance between $P_{F}(.| s)$ and $P_{F}(.| s')$ is:
\begin{equation}
    \text{OT}\left ( P_F(\cdot| s),P_F(\cdot| s') \right ) =\min _{\pi \in \prod\left(P_F(\cdot| s),P_F(\cdot| s')\right)}\langle \mathbf{C},\pi\rangle = \langle \mathbf{C},\pi^*\rangle.
\end{equation}
Specially, in EB-GFN experiments, all children $u_i$ of $s$ is a terminal state so $d(u_i,s_f) =-\log(1) = 0$. This makes $\mathbf{C} = 0$ and $\text{OT}\left ( P_F(\cdot| s),P_F(\cdot| s') \right ) = 0$. In Hyper-grid environment experiment, for terminal sate $s'$ because $c'_i =0$, we have:
\begin{align}
    \text{OT}\left ( P_F(\cdot| s),P_F(\cdot| s') \right ) =  \mathbf{H}(P_{F}(\cdot| s),P^{*}_{B}(\cdot| s)) - \log(P_{F}(s'|s)) & \nonumber \\
    & \hspace{- 8 em} +P_F(s'|s).(\log( P_B(s'|s))+\log(P_F(s'|s))).
\end{align}

The Hyper-grid environment ~\cite{DBLP:journals/corr/abs-2106-04399} (in Section \ref{ex: Hyper-grid environment}) and EB-GFN experiments~\cite{pmlr-v162-zhang22v} (in Section \ref{exp: Synthetic discrete probabilistic modeling tasks}) satisfy two condition in Theorem \ref{theorem: close form}.
In Biological Sequence Design~\cite{Jain2022BiologicalSD} (in Section \ref{exp: Biological sequence design })
such as protein and DNA sequences, the action space consists of actions adding a nucleic acid in $\{A,T,G,U\}$ and a amino acid respectively. Such settings satisfy former condition $a_i \neq a_k+a_h \quad \forall a_i,a_k,a_h \in \mathcal{A}
$. However, the later condition
$a_i+a_h = a_m+a_n, a_i \neq a_m \Longleftrightarrow a_i = a_n , a_h = a_m, a_i \neq a_m$ is no longer true because the order property of action space, i.e, $a_i+a_j \neq a_j+a_i$. In this situation, the third terms in Eqn.~\eqref{theorem: eq closeform} is zero and we can still using the formulation in Eqn.~\eqref{theorem: eq closeform}. Generally, the action space is independence and unique factorization.

\subsection{
Proof of Theorem \ref{theorem: dropout OT}} \label{proof:dropout OT }
By definition of the edge flow we have
\begin{equation}
    \sum _{\tau : s\to s' \in \tau}{P(\tau)}=\sum _{\tau : s\to s' \in \tau}{\frac{F(\tau)}{Z}}=\frac{F(s\to s')}{Z}=P(s \to s').
\end{equation}
From that equation, we find that
\begin{equation}
\begin{aligned}
    \mathbb{E}_{\tau \sim \pi _{\boldsymbol{\theta}}}( \mathcal{L}_{\text{OT}}(\tau)) 
    &= \mathbb{E}_{\tau \sim \pi _{\boldsymbol{\theta}}}\left(\sum _{s\to s' \in \tau}{\text{OT}\left ( P_F(\cdot|s),P_F(\cdot|s') \right )}\right) 
    %\\
    %&= \sum_{\tau}\left(\sum _{s\to s' \in \tau}{\text{OT}\left ( P_F(\cdot|s),P_F(\cdot|s') \right )}\right).P(\tau) 
    \\
    &= \sum_{\tau}\sum _{s\to s' \in \tau}{\text{OT}\left ( P_F(\cdot|s),P_F(\cdot|s') \right )}.P(\tau) 
    \\
    &= \sum_{s\to s'}\sum_{\tau: s\to s' \in \tau}{\text{OT}\left ( P_F(\cdot|s),P_F(\cdot|s') \right )}.P(\tau)
    \\
    &= \sum_{s\to s'}{\text{OT}\left ( P_F(\cdot|s),P_F(\cdot|s') \right )}\sum_{\tau: s\to s' \in \tau}{P(\tau)} 
    \\
    &= \sum_{s\to s'}{\text{OT}\left ( P_F(\cdot|s),P_F(\cdot|s') \right )}.P(s \to s')
    \\
    &\propto  \mathbb{E}_{s\to s' \sim  \pi _{\boldsymbol{\theta}}}( \text{OT}\left ( P_F(\cdot| s),P_F(\cdot| s') \right )).
\end{aligned}
\end{equation}
As a consequence, we obtain the conclusion of the theorem.
%({\color{blue} Need to indicate the location of these experiments in the paper.} 

\section{Experiment Settings}
In this part, we report experiment settings, including evaluation metrics for comparing the methods, hyper-parameter choices, and neural network architectures for all experiments. For biological sequence design tasks, we also give more details about the task description and datasets used for training.
\label{sec:appendix_experiments}

\subsection{Hyper-grid environment} \label{appendix:hypergrid}
\subsubsection{Evaluation criteria}
To evaluate the performance, we measure the KL divergence between the actual and empirical distribution of the last $2 \times 10^5$ visited states. The number of modes found during the training progress is also used to measure the learned models' performance. 

\subsubsection{Implementation details}

\textbf{GFlowNet: } For the implementation of the GFlowNet model, we also follow the framework of \cite{Malkin2022TrajectoryBI}: an MLP with two hidden layers of 256 dimensions each. The GFlowNet policy model, which includes both $P_F$ and $P_B$, is trained with a learning rate of $0.001$ while the learning rate for total flow $Z_{\theta}$ is $0.1$. We use a mini-batch size of $16$ and $62500$ training steps with the trajectory balance objective. 

\textbf{Proposed OT regularization} The regularization coefficient is $0.02$ for both Min OT, UB OT, and Max OT in our $4-D$ and $7-D$ hypergrid environment.

\subsection{Biological sequence design} \label{appendix:bioseq}
\subsubsection{Task description \& Datasets}
These experiments simulate the process of designing biological sequences, such as anti-microbial peptides, DNA, and protein sequences..., in drug discovery applications. This process often consists of an active loop with several rounds of ideating molecules and multiple-stage evaluations for filtering candidates, with rising levels of precision and cost. This characteristic makes the diversity of proposed candidates a considerable concern in the ideation phase because many similar candidates can all fail in the later phases.

Specifically, we consider the problem of finding objects $x$ in the space of discrete objects $\mathcal{X}$, that maximize a given oracle $f: \mathcal{X} \mapsto \mathbb{R}^{+}$. Here, we can only query this oracle $N$ times, each with an input batch of fixed size $b$. This can form $N$ rounds of evaluation in the active learning setting, where the generative policy is initially given a dataset $D_{0}=\left\{\left(x_{1}^{0}, y_{1}^{0}\right), \ldots,\left(x_{n}^{0}, y_{n}^{0}\right)\right\}$  collected from the oracle, where $y_{i}^{0}=f(x_{i}^{0})$ for $1 \leq i \leq n$. 

Because the oracle can only be called limited, we also train a supervised proxy model $M$ that predicts $y$ from $x$ to approximate the oracle $f$. Specifically, in $i$-th round, given the current dataset $\mathcal{D}_{i}$, this proxy model can be used as a reward function $R$ to collect additional observations to train our generative policy to propose a batch of candidates $\mathcal{B}_{i}=\left\{x_{1}^{i}, \ldots, x_{b}^{i}\right\}$. Then the current dataset $\mathcal{D}_{i}$ is updated for the next round of evaluation as $\mathcal{D}_{i+1}=\mathcal{D}_{i} \cup \left\{\left(x_{1}^{i}, y_{1}^{i}\right), \ldots,\left(x_{b}^{i}, y_{b}^{i}\right)\right\}$ where $y_{j}^{i}=f\left(x_{j}^{i}\right)$.

Following the framework of \cite{Jain2022BiologicalSD}, we will conduct experiments on the biological sequence design tasks:

\textbf{Anti-Microbial Peptide Design:} This task aims to generate short amino-acid sequences of length lower than $51$, which have anti-microbial properties. The vocabulary has $20$ amino-acids $[A,C,D,E,F,G,H,I,K,L,M,N,P,Q,R,S,T,V,W,Y]$. The active learning algorithm is evaluated for $N=10$ rounds, with the number of candidates generated each round $b=1000$. The initial dataset $\mathcal{D}_{0}$ contains $3219$ AMPs and $4611$ non-AMP sequences, which is collected from the DBAASP database \cite{DBLP:journals/nar/PirtskhalavaAGC21}. 

\textbf{TFBind 8: } The goal of this task is to generate DNA sequences of length $8$, which have high binding activity with human transcription factors. The vocabulary has $4$ nucleobases $[A,C,T,G]$. The active learning algorithm is evaluated for $N=10$ round, with the number of candidates generated each round $b=128$. The initial dataset $\mathcal{D}_{0}$ contains $32,898$ samples, which is half of all possible DNA sequences of length 8 having lower scores. The data and the oracle used are from \cite{doi:10.1126/science.aad2257}. 

\textbf{GFP: } The objective of this task is to generate protein sequences of length $237$ that have high fluorescence. The vocabulary is similar to the one of the AMP task (size $20$). The active learning algorithm is evaluated for $N=10$ round, with the number of candidates generated each round $b=128$. The initial dataset $\mathcal{D}_{0}$ contains $5,000$ samples, which is from \cite{Rao2019EvaluatingPT, Sarkisyan2016LocalFL} together with the oracle.

\subsubsection{Evaluation criteria}
To evaluate the performance, we also use the metrics as in \cite{Jain2022BiologicalSD}. Specifically, considering a set of candidates $\mathcal{D}$, we have the following metrics:

\textbf{Performance score: } mean score of the candidates in the set
    \begin{equation}
        \begin{aligned}
            \operatorname{Mean}(\mathcal{D})=\frac{\sum_{\left(x_{i}, y_{i}\right) \in \mathcal{D}} y_{i}}{|\mathcal{D}|},
        \end{aligned}
    \end{equation}

\textbf{Diversity: } a measurement of how well the generated candidates can capture the modes of the distribution implied by the oracle
    \begin{equation}
        \begin{aligned}
            \operatorname{Diversity}(\mathcal{D})=\frac{\sum_{\left(x_{i}, y_{i}\right) \in \mathcal{D}} \sum_{\left(x_{j}, y_{j}\right) \in \mathcal{D} \backslash\left\{\left(x_{i}, y_{i}\right)\right\}} d\left(x_{i}, x_{j}\right)}{|\mathcal{D}|(|\mathcal{D}|-1)},
        \end{aligned}
    \end{equation}
    where $d$ is a distance defined over $\mathcal{X}$, such as Levenshtein distance \cite{10.5555/1822502}. 

\textbf{Novelty: } a measure of the difference between the candidates in $\mathcal{D}$ and $\mathcal{D}_{0}$
    \begin{equation}
        \begin{aligned}
            \operatorname{Novelty}(\mathcal{D})=\frac{\sum_{\left(x_{i}, y_{i}\right) \in \mathcal{D}} \min _{s_{j} \in \mathcal{D}_{0}} d\left(x_{i}, s_{j}\right)}{|\mathcal{D}|}.
        \end{aligned}
    \end{equation}

These metrics will be evaluated on the set of candidates that have top $K$ scores $\mathcal{D}=\operatorname{TopK}\left(\mathcal{D}_{N} \backslash \mathcal{D}_{0}\right)$. 

\subsubsection{Implementation details}
For the implementation of the GFlowNet-AL baseline model,  we use the previously published implementation with slight changes, which follows the training setups of \cite{Jain2022BiologicalSD}:

\textbf{Proxy model}: We parameterize it as an MLP with two hidden layers, each having $2048$ hidden units, and use ReLU activation. We also use ensembles of 5 models with same architecture for uncertainty estimation. For the acquisition function, we use UCB $(\mu+\kappa \sigma)$ with $\kappa=0.1$. The proxy is trained with MSE loss using mini-batch of size $256$ and Adam optimizer with $\left(\beta_{0}, \beta_{1}\right)=(0.9,0.999)$ and learning rate $10^{-4}$. During training, early stopping is also used by evaluating the validation set containing $10\%$ of the data.  

\textbf{GFlowNet generator}: We use an MLP with 2 hidden layers of $2048$ hidden units each. The model is trained with trajectory balance objective as the main loss function, by using Adam optimizer with $\left(\beta_{0}, \beta_{1}\right)=(0.9,0.999)$. Additionally, $\log Z$ is trained with a learning rate of $10^{-3}$ for AMP, TF Bind 8 task, and $5 \times 10^{-3}$ for GFP task. Other hyper-parameters are shown in the following table:

\begin{table}[!ht]
\begin{center}
\begin{tabular}{lllll}
\toprule
 \textbf{Hyper-parameter} & \textbf{AMP} & \textbf{TF Bind 8} & \textbf{GFP}\\
\midrule
$\delta$ : Uniform Policy Coefficient  & 0.001 & 0.001 & 0.05\\
Learning rate & $5 \times 10^{-4}$ & $10^{-5}$ & $10^{-3}$\\
$m$ : Minibatch size & 32 & 32 & 32\\
$\beta$ : Reward Exponent $R(x)^{\beta}$ & 3 & 3 & 3\\
T : Training steps & 10,000 & 5,000 &50,000\\
\bottomrule
\end{tabular}
\label{table:hyperparameters}
\end{center}
\vspace{-0.15in}
\caption{Hyper-parameters for the GFlowNet.}
\end{table}

There are some changes in hyper-parameter choices and the number of active learning rounds in the TF Bind 8 task and the GFP task compared to the original training setups of \cite{Jain2022BiologicalSD}. However, during the experiment, we observed that these settings helped us get the closest results to the reported one in \cite{Jain2022BiologicalSD}.

\textbf{Proposed OT regularization} The regularization coefficients for Min OT, UB OT, and Max OT are the same for each biological sequence design task. Specifically, the coefficients for the AMP, TF Bind 8, and GFP task are $0.025$, $0.1$, and $0.02$ correspondingly.

\subsection{Synthetic discrete probabilistic modeling tasks} \label{appendix:discrete_modelling}

\subsubsection{Evaluation criteria}
To evaluate the performance, we keep the same evaluation criteria in \cite{pmlr-v162-zhang22v}, where they use the NLL of a large independent sample of ground truth data and the exponential Hamming MMD (\cite{JMLR:v13:gretton12a}) between ground truth data and
generated samples as performance metrics. To measure NNL and MMD, we use 10 fixed sets, and each set consists of 4000 ground truth data samples.  

\subsubsection{Implementation details}

\textbf{GFlowNet: } For the implementation of the GFlowNet model, we use an MLP with 2 hidden layers of 512 dimensions each. The GFlowNet policy model, which includes both $P_F$ and $P_B$, is trained with a learning rate of $0.001$. We use a mini-batch size of $128$ and $1e5$ training steps with the trajectory balance objective. 

\textbf{EBMs: } For the implementation of the Energy-Based Model, we use an MLP with 3 hidden layers of 256 dimensions each. The learning rate is $0.001$.

\textbf{Proposed OT regularization}: The regularization coefficient is $0.001$ for both Min OT and UB OT and is the same for all tasks.

\end{document}